\documentclass[twoside]{article}

\usepackage[accepted]{aistats2025}

\usepackage[utf8]{inputenc}    
\usepackage[T1]{fontenc}       
\usepackage{hyperref}          
\usepackage{url}               
\usepackage{booktabs}          
\usepackage{amsfonts}          
\usepackage{nicefrac}          
\usepackage{microtype}         
\usepackage{xcolor}            

\usepackage{natbib}
\usepackage{graphicx}
\usepackage{subfigure}
\usepackage{mathrsfs}
\usepackage{multirow}
\usepackage{array}
\usepackage{algorithm}
\usepackage{algorithmic}
\usepackage{wrapfig}

\begin{document}

\twocolumn[

\runningtitle{Partial Information Decomposition for Data Interpretability}

\aistatstitle{Partial Information Decomposition for\\ Data Interpretability and Feature Selection}

\aistatsauthor{
  Charles Westphal \And
  Stephen Hailes \And
  Mirco Musolesi
}

\aistatsaddress{
  University College London \\ \texttt{charles.westphal.21@ucl.ac.uk} \And
  University College London \\ \texttt{s.hailes@ucl.ac.uk} \And
  University College London \\ 
  University of Bologna \\ \texttt{m.musolesi@ucl.ac.uk}
}


]

\begin{abstract}
  In this paper, we introduce Partial Information Decomposition of Features (PIDF), a new paradigm for simultaneous data interpretability and feature selection. Contrary to traditional methods that assign a single importance value, our approach is based on three metrics per feature: the mutual information shared with the target variable, the feature’s contribution to synergistic information, and the amount of this information that is redundant. 
In particular, we develop a novel procedure based on these three metrics, which reveals not only how features are correlated with the target but also the additional and overlapping information provided by considering them in combination with other features. We extensively evaluate PIDF using both synthetic and real-world data, demonstrating its potential applications and effectiveness, by considering case studies from genetics and neuroscience.
\end{abstract}

\section{INTRODUCTION}
\label{intro}

Data interpretability and feature selection are key areas of research in machine learning (ML). One way in which researchers quantify the relative significance of a model’s input attributes is by assigning each of them a feature importance. Multiple paradigms for deriving feature importance have been developed over the past decades. Some of these, inspired by techniques from economics, are based on \textit{Shapley values} \citep{Keinan2004,cohen2007,Apley2020,debeer20,catav21,kwon22,janssen2023}; others instead rely on the quantification of the correlation between the model’s target and its features \citep{peng05,brown12,gao16,chen18,schnapp21,covert23}. Feature importance can be used to explain the relationships within the underlying data and select features \citep{covert2020,weinberger2024feature}. However, the authors \citet{catav21} and \citet{janssen2023} highlighted inconsistencies in achieving these dual objectives simultaneously. To emphasize this, they considered two perfectly correlated features. If selecting from them, an optimal method would assign one of the two as important, discarding the other. On the other hand, for an exact understanding of how they reduce the uncertainty of the target, they should be assigned equal importance.

\begin{figure*}[t]
\vskip 0.2in
\begin{center}
\centerline{\includegraphics[width=\textwidth]{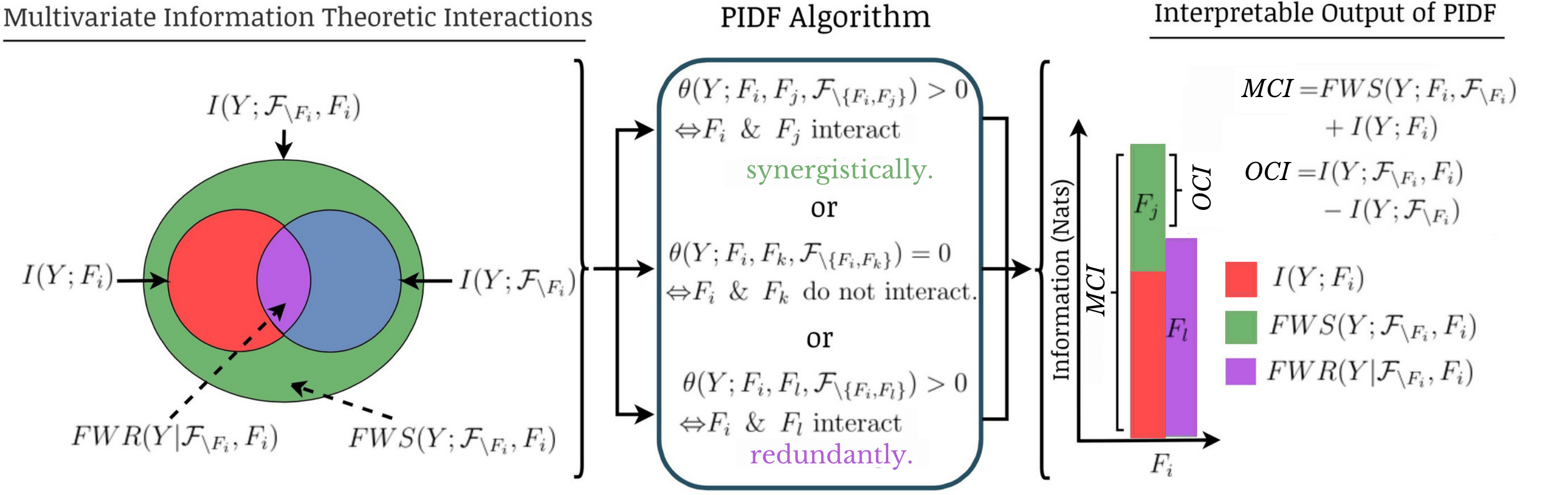}}
\caption{PIDF at a glance. The diagram on the left shows the interactions that characterize how feature $F_i$ interacts with the remaining features $(F_j, F_k)$ to describe the target. The bar graph on the right shows how PIDF can be used to represent these quantities in an interpretable manner.}
\label{fig:ven}
\end{center}
\vskip -0.2in
\end{figure*}

In this paper, we develop \textit{partial information decomposition of features (PIDF)}, a novel method that simultaneously explains data and selects features, even in the presence of complex interactions such as those highlighted by \citet{catav21} and \citet{janssen2023}. PIDF relies on novel definitions of synergy, redundancy, and mutual information \textit{per feature}\footnote{Code available at: \url{https://github.com/c-s-westphal/PIDF}.}. PIDF builds upon partial information decomposition (PID) \citep{chechik2001group,williams2010}, the classic framework for studying variable interactions.
However, unfortunately, PID is based on the calculation of a large number of intractable quantities \citep{makkeh2019maxent3d_pid}, making it essentially inapplicable to real-world datasets. Instead, we elect to use the interaction information (II) to detect synergy and redundancy. However, II is designed to be applied to generic sets of variables and does not possess the ability to differentiate between synergy and redundancy on a per-feature basis \citep{williams2010}. In response to this, we introduce the concepts of \textit{feature-wise synergy (FWS)} and \textit{feature-wise redundancy (FWR)}, which are not only able to capture these relationships but are also computationally tractable even when describing large datasets.
PIDF isolates the FWR, FWS and MI per feature and allows them to be presented in an interpretable manner as illustrated in Figure \ref{fig:ven}. We show that PIDF effectively explains the data and selects optimal features. We evaluate PIDF's versatility and performance on various synthetic and real-world datasets, demonstrating its robustness and efficacy.

\section{RELATED WORK}\label{sec:rel_work}
\textbf{Dataset interpretability.}
Feature importance has been used to enhance interpretability and select features \citep{fisher2019all,covert2020,aas21}. However, \citet{catav21} highlighted that these dual perspectives do not appear universally consistent. This dilemma prompted \citet{catav21} to propose axioms for assigning feature importance when the goal was solely to explain the underlying data. Building on these principles, \citet{janssen2023} further refined Catav's axioms for interpretability, while also enhancing speed and ensuring unrelated variables do not receive a non-zero importance rating. That said, these axioms fail to handle complex redundancies, consider the following two cases. First, suppose there exist three features, two of which are indistinguishable, one is unique, and all three reduce the uncertainty of the target equally.
%
According to both the `Invariance under Redundant Information and Symmetry under Duplication'  and `Elimination'  axioms presented in \citet{janssen2023} and \citet{catav21} all three variables will be assigned the same importance. We contend that to explain the underlying relationships within data, it is essential to identify when information is redundant and consequently more disposable. 
Secondly, while these axioms of feature importance 
do account for higher-order interactions, they cannot distinguish when the interactions are synergistic and which features are involved. 


\textbf{Partial information decomposition of features for feature selection.}
Within the context of feature selection, the most relevant work to ours is that produced by \citet{wollstadt23} as they truly develop a framework for the PID of features. Here, they are not using their PID adaptation to describe the relationships within data and select features. Rather, it is used to explain why certain features were chosen over others. Their selection method is a derivative of the conditional likelihood maximization framework unified by \citet{brown12}; in which a feature is selected if augmenting it to the set of those already chosen maximizes the correlation with the target. At each round of selection, they decompose the contributions of each potential feature according to their adaptation of the PID paradigm. Unfortunately, this method is unable to identify fully synergistic variables (i.e, variables that considered alone are non-informative, but combine to describe the target). Such features will not be selected using this method, and their synergy will remain undiscovered. Another feature selection method built upon the theoretical foundations of PID is developed in \citet{westphal24}. The authors study how entropy is transferred from the features to the target, sequentially eliminating those with a negligible contribution. This method effectively identifies synergistic variables; however, it does not consistently identify the smaller of two redundant feature subsets.  
Consider a scenario where a subset of features conveys the same information as a single variable. The method proposed in \citet{westphal24} falls short in consistently choosing the more parsimonious option, which in this case, would be a single variable. 

\section{RESEARCH QUESTIONS}

Following the discussion of the related work in the previous section, we can identify three open research questions for which we will demonstrate that PIDF represents an effective solution (in particular, the first two relates to interpretability and the final one to feature selection):

\begin{itemize}
\item \textit{The redundant variables question (RVQ):}
Is it possible to distinguish two fully redundant features from one of their non-redundant counterparts if all three reduce the uncertainty of the target equally? 
\item \textit{The synergistic variables question (SVQ):} Let us consider the case where two features combine to be fully informative regarding the target, whereas, if taken singularly, they are completely uninformative. Is it possible to reveal these variables are only useful in combination? 
\item \textit{The minimal subsets question (MSQ):} 
Suppose there exist multiple perfectly redundant feature subsets, is it possible to consistently select the smallest one?
\end{itemize}
\section{PIDF AT A GLANCE}\label{sec:prelim}



\subsection{Overview}
Let the input to a ML model be denoted $x = (f_1, f_2,..., f_N)$, where $f_i$ is a feature instance. By random sampling from the elements of our input list $f_i$, and from the realizations of the ground truth $y$, we represent not only all possible features as a set of random variables (written as $\mathcal{F} = \{F_1, F_2\dots F_N\}$), but also the ground truth (written as $Y$). Using these distributions, PIDF aims to elucidate \textit{not only the importance of a feature, but also the inter-feature interactions}. 
PIDF derives three information-theoretic quantities per feature, namely FWS, FWR, and MI. These quantities are then used to understand how the feature of interest (labeled $F_i \in \mathcal{F}$) interacts with the remaining features (labeled $F_j, F_k, F_l \in \mathcal{F}_{\backslash F_i}$) to describe the target. By considering this fine-grained decomposition and analysis, we are also able to deal with limit cases, such as the perfectly correlated features problem highlighted by \citet{catav21} and \citet{janssen2023}.

To enhance the interpretability of PIDF's results, we present these three quantities as shown in Figure \ref{fig:ven}. Per feature we will have three bars, the first of which (red), will describe the MI shared with the target. The second (green), will quantify the FWS this feature provides in combination with the remaining features. These red and green bars will stack to give the \textit{maximal conditional information} (MCI). To the right-hand side of this, we will present a purple bar that reveals the extent to which the MCI provided by this feature is redundant. FWS and FWR arise due to interactions; consequently, we label these bars with the other features with which these interactions occur. For instance, in Figure \ref{fig:ven}, we observe that $F_i$ interacts synergistically with $F_j$ and redundantly with $F_l$. 

\subsection{A Practical Example}
To provide a better intuition of the proposed approach, let us consider the following example. Suppose we are trying to predict housing prices in Northern California, using the following per-property data: longitude, latitude, and distance to the coast. Individually, given its geography, longitude, and latitude offer limited insight into housing prices. However, they combine synergistically to provide an exact location — a highly informative feature. Furthermore, longitude is redundant when compared to coastal proximity. 
The application of PID to this example would require  the estimation of 20 different terms, which is not practically possible with the existing techniques \citep{williams2010,makkeh2018broja,makkeh2019maxent3d_pid}.
On the other hand, tractable methods, such as \citet{chechik2001group,williams2010,Griffith2012,Bertschinger2012,griffith2015}, are unable to capture the interactions of individual features, instead suggesting that the combination of longitude, latitude, and distance to the coast provides overall redundant information. For this reason, we re-formulate both synergy and redundancy as FWS and FWR, respectively. These quantities can reveal detailed interactions on a per-feature basis, while remaining tractable. Leveraging these quantities, we develop PIDF, an algorithm whose interpretable representation of FWS, FWR, and MI, allows for simultaneous data interpretation and feature selection.

\section{FEATURE-WISE PID}\label{sec:fw_pid}
In this section, we introduce FWS and FWR more formally. First of all, we use the definition of MI as introduced in \citet{shannon1948}.
PID exhaustively describes all possible feature interactions in a set of variables \citep{williams2010}. Consequently, one may expect that FWS and FWR are derived from the PID terms. However, as discussed in the introduction, these values are far too numerous and their calculation too computationally-expensive to be considered applicable to real-world datasets\footnote{For a more detailed comparison of PIDF to PID please refer to Appendix \ref{app:comarison}.}. Consequently, the definitions FWS and FWR extend the concept of interaction information (II) \citep{williams2010,chechik2001group}. The II for two features is defined as follows: $I(Y;F_1;F_2) = I(Y;F_1,F_2) - I(Y;F_1) - I(Y;F_2)$, if these two features interact synergistically then $I(Y;F_1,F_2)>0$, if the opposite is true, this indicates redundancy. The simplest approach would be to use II in its current form to calculate both the redundancy and synergy components of our output (purple and green bars in Figure \ref{fig:ven}). However, this approach has considerable shortcomings as detailed in the following section.

\textbf{Measuring feature-wise synergy.}
When assessing the synergistic contribution of an individual feature, the aim is to determine how much extra information a feature can reveal about the target by combining with the remaining features. One method, adopted in \citet{wollstadt23}, rewrites the MI considering the feature set as a two-body system, where the feature of interest interacts with those remaining, such that: $\mathcal{F} = \{F_i, \mathcal{F}_{\backslash F_i}\}$. We re-apply this reasoning to the II and we obtain: $I(Y;F_i;\mathcal{F}_{\backslash F_i}) =  I(Y;F_i,\mathcal{F}_{\backslash F_i})  -  I(Y;\mathcal{F}_{\backslash F_i}) -  I(Y;F_i).$ However, this definition cannot describe features that interact both redundantly and synergistically. To explain this, we work through the following simple example: suppose that $F_i$ is completely uninformative in itself, but combines with $F_j \in \mathcal{F}_{\backslash F_i}$ via an XOR function to fully describe $Y$ (satisfying $I(Y;F_i,F_j) = H(Y)$). Moreover, let us suppose there also exists $F_k \in \mathcal{F}$ which is identical to $F_i$ ($F_i \equiv F_k$). In this case, the perfect redundancy between $F_i$ and $F_k$ ensures that $I(Y;F_i;\mathcal{F}_{\backslash F_i}) = 0$. However, for a full understanding of the underlying data we would like to know that the variable $F_i$ did provide synergistic information about $Y$ in combination with $F_j$, despite it being redundant. Motivated by this example, we define the FWS as the maximum extra information gained by adding a feature ($F_i$) to any subset of $\mathscr{P}(\mathcal{F})$. More formally, this becomes: $FWS(Y;F_i;\mathcal{F}_{\setminus F_i}) =  \max_{\mathcal{P}^{ms}\in \mathscr{P}(\mathcal{F}_{\backslash F_i})} I(Y;F_i;\mathcal{P}^{ms}).$ This describes the extra reduction in uncertainty of the target caused by considering $F_i$ collectively with $\mathcal{P}^{ms}$, \textit{the subset of maximum synergy}. Combining $FWS(Y;F_i;\mathcal{F}_{\setminus F_i})$ with $I(Y;F_i)$ results in the MCI ($MCI(Y;F_i,\mathcal{F}_{\backslash F_i}) = I(Y;F_i) + FWS(Y;F_i;\mathcal{F}_{\setminus F_i})$), i.e., the maximum possible reduction in uncertainty of the target caused by the feature $F_i$.

\textbf{Measuring feature-wise redundancy.}
When assessing feature-wise redundancy, our goal is to quantify to what extent the MCI ($FWS(Y;F_i;\mathcal{F}_{\setminus F_i})+I(Y;F_i)$) is redundant. Consequently, we require a value with the following properties: Firstly, it should be positive. Secondly, it should be smaller than the MCI. Consequently, we define the FWR as: $FWR(Y;F_i;\mathcal{F}_{\backslash F_i}) = FWS(Y;F_i;\mathcal{F}_{\backslash F_i})- I(Y; F_i; \mathcal{F}_{\backslash F_i}) $. This expression quantifies the discrepancy between the maximum and overall reduction in uncertainty of the target variable $Y$ caused by feature $F_i$ when combined with any subset of features and with all remaining features, respectively. Intuitively, such a definition aligns with standard definitions of redundancy, that measure the difference between the maximum and true entropy of a variable ensemble \citep{Plaus1996}. Moreover, FWR satisfies the desirable properties outlined to begin this section and can be related to the MCI via the following theorem:

\noindent \textbf{Theorem 1:} \textit{The difference between the MCI (i.e., $FWS(Y;F_i;\mathcal{F}_{\setminus F_i})+I(Y;F_i)$) and the $FWR(Y;F_i;\mathcal{F}_{\backslash F_i})$ is the overall conditional information (OCI) contributed by $F_i$. More formally: $FWS(Y;F_i;\mathcal{F}_{\setminus F_i})+I(Y;F_i)-FWR(Y;F_i;\mathcal{F}_{\backslash F_i}) =I(Y;\mathcal{F}_{\backslash F_i},F_i) - I(Y;\mathcal{F}_{\backslash F_i}).$}

 \noindent \textit{Proof.} See Appendix \ref{app:proof_tint}. \\ \noindent In Theorem 1, we have shown that the difference between the MCI, and our definition of feature-wise redundancy $FWR(Y;F_i;\mathcal{F}_{\backslash F_i})$, is the non-redundant information gained by adding $F_i$ to the set $\mathcal{F}_{\backslash F_i}$ (labeled OCI in Figure \ref{fig:ven}). It follows that we have developed a consistent theory of feature-wise PID. 

\section{PIDF}
Thus far, we have introduced FWS and FWR, explaining that we aim to isolate these quantities alongside MI for data interpretability. The simplest method to do this would be to maximize $I(Y;F_i;\mathcal{F}_{\backslash \{F_i\}})$ across all possible subsets to find $\mathcal{P}^{ms}$  for each feature $F_i$. However, this is computationally inefficient. To overcome expensive subset searches, we perform the following steps. We first introduce a measure with the ability to evaluate pairwise synergy or redundancy. This allows us to systematically assess whether feature $F_j$ should be added to $F_i$'s set of maximum synergy, avoiding costly subset searches. We then discuss potential applications of this measure and highlight some flaws. Before using these examples to motivate an assumption about the data that bounds the values our measure can take. Finally, we leverage these bounds for a practical implementation of PIDF.

\subsection{Interaction Information Based Measure} To begin, we introduce the following measure: \begin{align}
\theta(Y; F_i; F_j; \mathcal{P}_{\setminus \{F_i, F_j\}}) &= I(Y; F_i; \mathcal{P}_{\setminus \{F_i\}}) \nonumber \\
&\quad - I(Y; F_i; \mathcal{P}_{\setminus \{F_i, F_j\}}),
\end{align}
where $\mathcal{P}_{\setminus \{F_i, F_j\}} \in \mathscr{P}(\mathcal{F}_{\setminus \{F_i, F_j\}})$ and $\mathcal{P}_{\setminus \{F_i\}} \in \mathscr{P}(\mathcal{F}_{\setminus \{F_i\}})$. By comparing two IIs, $\theta(Y;F_i;F_j;\mathcal{P}_{\backslash \{F_i, F_j\}})$ quantifies how feature $F_j$ affects $F_i$'s ability to reduce the uncertainty of the target. Positive values of $\theta$ indicate that feature $F_j$ interacts synergistically with $F_i$, and should be added to the $F_i$'s set of maximum synergy. Meanwhile, negative values imply redundant information is shared between the two features. Using $\theta$ we aim to illuminate all synergistic interactions, enabling the calculation of FWS and FWR.

\subsection{Handling Redundant Information} We now describe how, if features in the set $\mathcal{F}_{\backslash F_i}$ provide redundant information regarding feature $F_i$, it impedes the ability to calculate $FWS(Y;F_i;\mathcal{F}_{\setminus F_i})$ and $FWR(Y;F_i;\mathcal{F}_{\backslash F_i})$. To do this, we first point out that $\theta(Y; F_i; F_j; \mathcal{P}_{\setminus \{F_i, F_j\}}) $,  if positive, indicates that feature $F_j$ can be added to $F_i$'s set of maximum synergy. For a simple and intuitive example of the application of this measure, suppose we calculate $\theta(Y; F_i; F_j; \mathcal{F}_{\setminus \{F_i, F_j\}}) \forall j$, adding those with positive values to $F_i$'s set of maximum synergy. However, this is potentially problematic. Suppose there exists another feature $F_k$ that is identical to feature $F_i$. Even if $F_i$ combines synergistically with other features, all values of $\theta(Y; F_i; F_j; \mathcal{F}_{\setminus \{F_i, F_j\}})$ will be equal to zero. To uncover synergistic interactions, we first need to remove redundant information from the set of all features. 

A potential solution would be to remove all features ($F_j, F_k \in \mathcal{F}_{\setminus F_i}$) with a non-negligible MI with $F_i$. However, this approach suffers from the following drawback. The MI between two features does not account for redundant interactions beyond pairwise relationships. For instance, multiple variables can interact synergistically to provide redundant information about $F_i$.
In practice, this is a limit case.
As a result, we assume that other features ($F_j \in \mathcal{F}_{\setminus F_i}$) do not combine synergistically ($\mathcal{P} \in \mathscr{P}(\mathcal{F}_{\setminus F_i})$) to provide redundant information about the feature of interest ($F_i$). 

\textbf{Assumption 1:} \textit{Features cannot combine synergistically to provide redundant information about other features.
More formally: $I(F_i;\mathcal{P}_{\setminus F_j}) + I(F_i;F_j) \geq I(F_i;\mathcal{P})$.}

Under this assumption, redundancy arises solely from pairwise MI between features, greatly simplifying the search for redundant features. We now show that, given this assumption, $I(F_i; F_j)$ serves as an indicator of whether $F_j$ is redundant or synergistic with respect to $F_i$.

\textbf{Theorem 2:} \textit{The upper and lower bounds of $\theta(Y;F_i;F_j;\mathcal{P}_{\backslash \{F_i, F_j\}})$ are a function of the pairwise MI between $F_i$ and $F_j$ ($I(F_i; F_j)$).
More formally: $2H(F_i)- I(F_i; F_j) \geq \theta(Y;F_i;F_j;\mathcal{P}_{\backslash \{F_i, F_j\}}) \geq - I(F_i; F_j).$}

\textit{Proof.} See Appendix \ref{app:proof_t2}.  \\ The above states that the lower bound of our measure is $-I(F_i;F_j)$. Given that if $\theta(Y;F_i;F_j;\mathcal{P}_{\backslash \{F_i, F_j\}}) < 0$ then $F_j$ is redundant with respect to $F_i$, it is reasonable to use $I(F_i;F_j)$ as an indicator of redundancy. Meanwhile, the upper bound is also a function of $-I(F_i;F_j)$. This indicates the opposite, i.e., low values of $I(F_i;F_j)$ imply that $F_j$ is more likely to interact synergistically with $F_i$. 

As introduced in Section \ref{sec:fw_pid}, deriving FWS and FWR requires identifying the set of maximum synergy, $\mathcal{P}^{ms}$, which includes all features, $F_j$, that have a positive value of $\theta(Y; F_i; F_j; \mathcal{P}_{\backslash {F_i, F_j}})$. To achieve this, it is necessary to identify and eliminate all redundant elements. Conveniently, in this section, we have shown that $I(F_i; F_j)$ is an indicator of redundancy. Combining these observations forms the basis of our straightforward implementation of PIDF, where we first order the features in descending order of $I(F_i; F_j)$ (from most to least redundant) and then remove those deemed redundant according to $\theta(Y; F_i; F_j; \mathcal{P}_{\backslash {F_i, F_j}})$\footnote{The reason that we still evaluate $\theta$ for all $F_j$, even for those features with positive values of $I(F_i;F_j)$, is because a non-zero MI between two features does not completely rule out the possibility that they interact synergistically in the presence of other variables. }. Via this assumption and simple implementation, we can calculate FWS and FWR without costly subset searches. We now describe PIDF in detail.

\subsection{PIDF in Practice}\label{sec:pip}
\textbf{Description of PIDF.} For all possible $F_i$, repeat the following steps: firstly, calculate $I(Y;F_i)$ to appear as the red bar in Figure \ref{fig:ven}. Then, rank the remaining features $F_j \in \mathcal{F}_{\setminus F_i}$ in descending order of $I(F_i, F_j)$ as, according to our assumption, this is an indicator of redundancy. Thirdly, iterate through the ranked features checking if they are truly redundant according to $\theta(Y;F_i;F_j;\mathcal{F}_{\backslash \{F_i, F_j\}})$ (the box titled `PIDF Algorithm' in Figure \ref{fig:ven} illustrates this process).  If so, they can be removed from the set $\mathcal{F}_{\backslash F_i}$. We continue to remove such features until the remainder are no longer redundant, indicating the identification of the set of maximum synergy for the feature $F_i$. By completing these steps for all $F_i \in \mathcal{F}$ we derive the FWS, FWR, and MI for each feature as desired. For an algorithmic definition of this process, refer to Algorithm \ref{alg:main} in Appendix \ref{app:algs}, and, for an example calculation, refer to Appendix \ref{app:example_calc}. 

\textbf{Adapting PIDF for varying estimates.} II, FWS, and FWR can be directly derived from MI. That said, unless restricting oneself to discreet data, exact calculations of MI are not possible. Consequently, we estimate it using methods outlined by \citet{belghazi2018} (we motivate this choice in Appendix \ref{app:mi_est}). These estimates can randomly fluctuate, which can lead to the erroneous identification of redundant features during the evaluation of line 7 in Algorithm \ref{alg:main}. To overcome this, when estimating any MI, the process is repeated five times. For feature $F_j$ to then be considered redundant with respect to the chosen feature $F_i$ there has to be a 95\% certainty that $\theta(Y;F_i;F_j;\mathcal{F}_{\backslash \{F_i, F_j\}}) < 0$. Otherwise, the feature is classed as non-redundant. 

\textbf{Interpretability of FWR.} We assume that features can provide redundant information regarding one another, provided that it is not synergistic. This implies that the total redundant information in the features must be a linear combination of individual contributions. Consequently, we need not present redundancy as just one bar, but rather as a stack of all the individual contributions from all the redundant features. Please refer to Appendix \ref{app:proof_fwr} for a formal proof of this statement.
As a concrete example, consider the diagram labeled as MSQ in Figure \ref{fig:novsynth}. For feature $F_0$, the FWR on the right-hand side of the MCI is not represented by a single bar, but rather as one composed by multiple stacked redundant contributions from individual features. The possibility of discriminating between the different redundant contributions through PIDF allows for better overall interpretability, by highlighting which features are the most redundant.

\subsection{PIDF for Feature Selection}\label{sec:fs}
We now provide a procedure for feature selection based on the output of PIDF. First of all, we select all features with a non-redundant contribution, i.e., any features whose MCI exceeds their FWR. We then rank the remaining features by their MCI and check in descending order if any features with which they had a non-negligible MI have already been selected. If so, we do not add this feature; otherwise, we proceed to add it. As was the case for PIDF, exceeding thresholds must be done with a minimum of $95\%$ certainty, where the MIs at the basis of these calculations have been estimated using multiple random seeds. The steps of the procedure are detailed in Algorithm \ref{alg:fs} (Appendix \ref{app:algs}). For a fully worked example of these steps, please refer to Appendix \ref{app:example_calc}.

\section{EXPERIMENTAL EVALUATION}
\subsection{PIDF for Data Interpretability}
We first evaluate PIDF’s ability to explain data, comparing it to three baseline methods: UMFI \citep{janssen2023}, MCI \citep{catav21} and PI \citep{breiman2001}. For a full discussion of the hyperparameters used for all methods please refer to Appendix \ref{app:hyper}. First, we provide a detailed explanation on how to interpret the results of PIDF. Following this, we present the remaining results, initially on synthetic data and subsequently on real-world data.

\subsubsection{Example of How to Interpret the Results of PIDF}\label{app:guided_exmaple}

\begin{figure*}[t] 
\begin{center}
\centerline{\includegraphics[width=\textwidth]{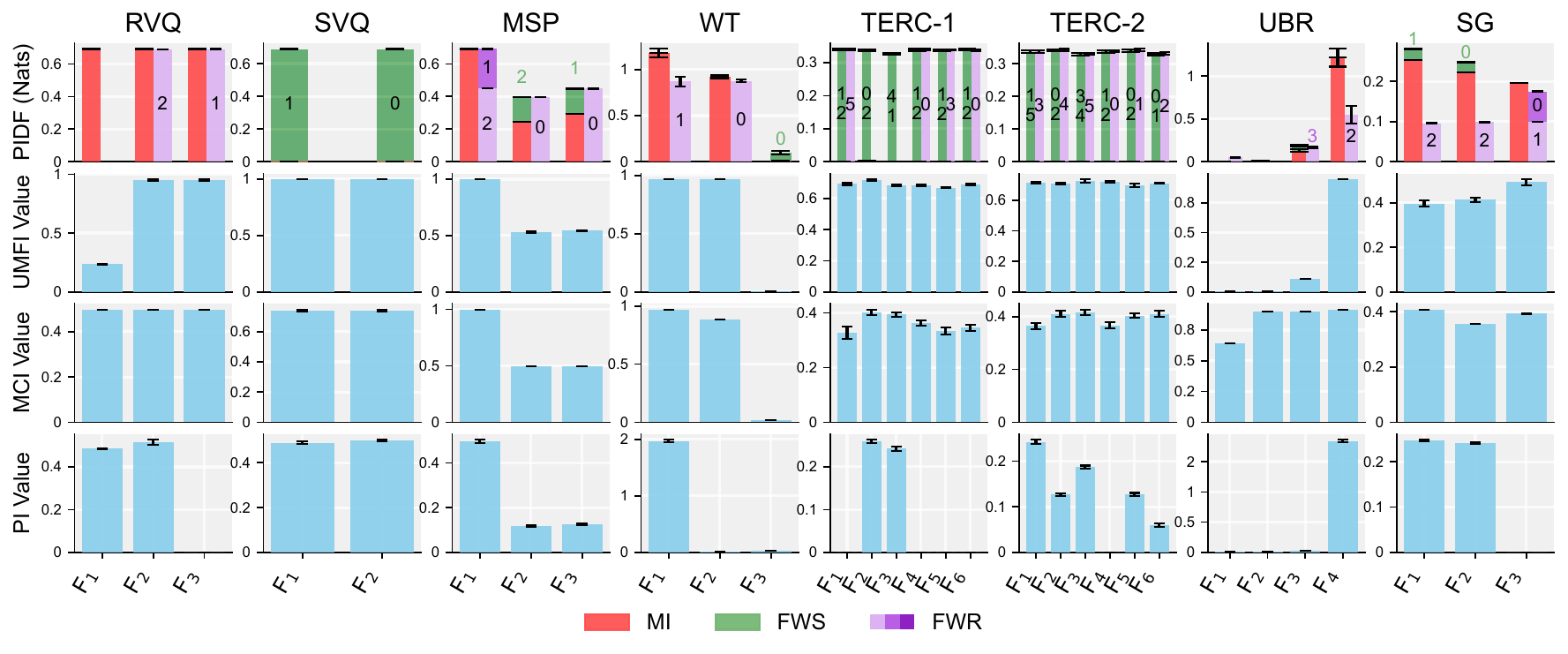}} 
\caption{Comparison of feature importance indicators using synthetic datasets.}
\label{fig:novsynth}
\end{center}
\vskip -0.2in
\end{figure*}

In this section, we discuss an example of how to interpret the results of PIDF, we refer to Figure \ref{fig:novsynth}. In this illustration, the top row represents the results of PIDF. We guide the reader through the Wollstadt toy example dataset \citep{wollstadt23}, as it is known to be characterized by both redundant and synergistic relationships (as shown in the graph titled WT). Firstly, we investigate the bars associated with $F_0$. We observe that $F_0$ shares a large MI with the target (red bar). However, it is also clear that the majority of this information is redundant with respect to $F_1$. We know the feature is redundant with respect to $F_1$ looking at the number (1) reported in $F_0$'s purple bar (i.e., the numbers in the bars represent the subscripts of the remaining features involved). If we now investigate the relationships associated with the feature $F_1$, we observe that this feature also provides significant information; however, it is entirely redundant with respect to $F_0$. Finally, investigating $F_2$, we see this final feature combines with $F_0$ to provide some non-redundant synergistic information (see the green 0 above the bar). Consequently, using Figure \ref{fig:novsynth}, we have identified the following relationships in the data:
\begin{enumerate}
    \item $F_0$ provides significant information; although, some of it is redundant with respect to $F_1$.
    \item  $F_1$ provides significant information; however, \textit{all} of it is redundant with respect to $F_0$ (this indicates that $F_0$ contains all the information in $F_1$ and some).
    \item  $F_2$ combines synergistically with $F_0$ to provide some non-redundant information \citep{wollstadt23}.
\end{enumerate} 
\begin{figure*}[h]
\begin{center}
\centerline{\includegraphics[width=\textwidth]{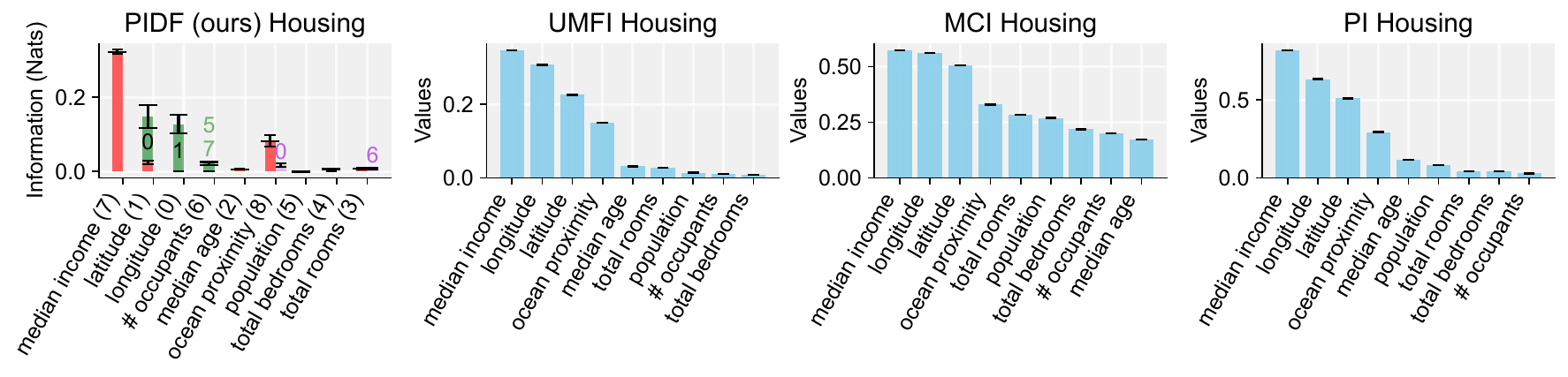}}
\caption{Comparison of feature importance indicators applied to the California housing dataset.}
\label{fig:housing}
\end{center}
\vskip -0.4in
\end{figure*}
\subsubsection{Synthetic Datasets} \label{sec:novsyn_expts} 

We now discuss the results obtained when applying PIDF to synthetic datasets. The first three of these datasets were designed to highlight how current methods fail to properly address the questions introduced in Section \ref{sec:rel_work}. The following five are taken from pre-existing works. A description of these datasets can be found in Appendix \ref{app:synth_desc}. 

\textbf{Redundant variables question (RVQ) dataset.} We now design an extremely simple dataset to demonstrate our method’s ability to answer the RVQ, as introduced in Section \ref{sec:rel_work}. The dataset is designed such that $F_1\equiv F_2$, meaning these two features are fully redundant with respect to one another, as revealed by PIDF in Figure \ref{fig:novsynth}. Meanwhile, using MCI one would assume that we have three features of equal importance, without knowledge of the redundancy.

\textbf{Synergistic variables question (SVQ) dataset.}
We now present the evaluation conducted to answer the SVQ introduced in Section \ref{sec:rel_work}. In this dataset, the target is formed by the synergistic combination of the two features. This is indicated by the subscripts in the synergy bar for features $F_0$ and $F_1$, as shown in Figure \ref{fig:novsynth}.
We observe that only PIDF reveals that $F_0$ combines with $F_1$ synergistically. 

 \textbf{Multiple subsets question (MSQ) dataset.} This dataset is characterized by redundancy between the subset $\{F_1,F_2\}$ and the variable $F_0$. In Figure \ref{fig:novsynth}, we observe that our method again resolves the complex relationships correctly.

 
 \textbf{TERC-1 \& TERC-2.} These datasets, introduced by \citet{westphal24}, are characterized by features that are simultaneously both redundant and synergistic, as illustrated in Figure \ref{fig:novsynth}. Moreover, through the labels of the green and purple bars, one can understand with which other features these interactions occur. 
\begin{figure*}[h]
\begin{center}
\centerline{\includegraphics[width=\textwidth]{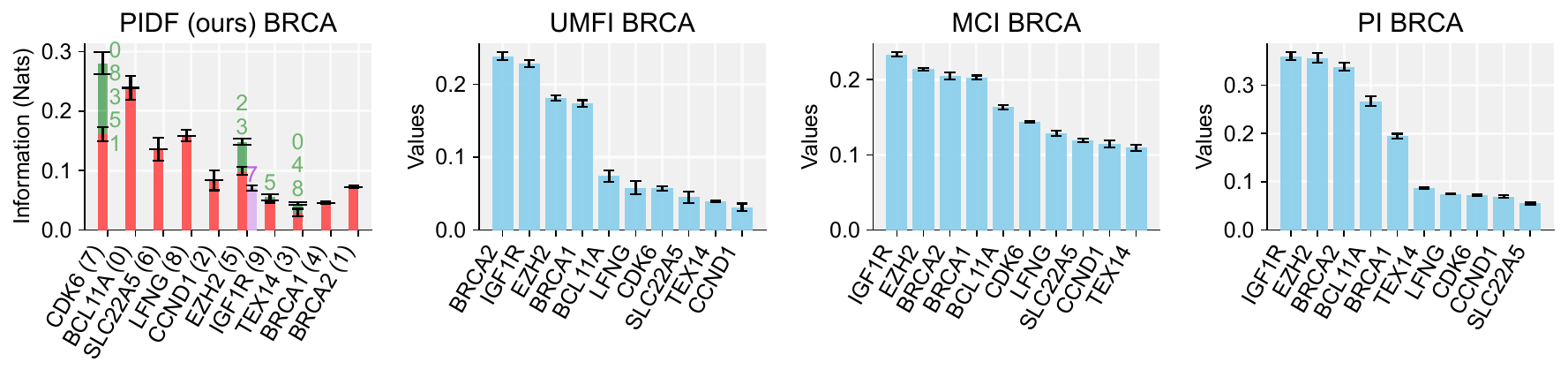}}
\caption{Gene importance in the BRCA dataset.}
\label{fig:brca}
\end{center}
\vskip -0.4in
\end{figure*}
\label{fig:genes} 

\textbf{UMFI blood relation dataset (UBR).} \citet{janssen2023} used the UBR dataset to demonstrate that their method only assigns non-negligible importance to features that are related to the target via a directed path or causal ancestor \citep{pearl1986}. Despite neither $F_1$ nor $F_2$ satisfying these conditions, the methods introduced in \citet{catav21} still assigned them as important. PIDF avoids these issues and, additionally, unlike UMFI, it reveals that the information provided in $F_2$ is redundant with respect to $F_3$. 

\textbf{Synthetic genes dataset (SG).} Our final synthetic dataset was developed in \citet{anastassiou07} to demonstrate that two genes could interact synergistically, while others are redundant. We observe that our method, similarly to the PID analysis in \citet{anastassiou07}, has the capability to reveal such relationships. However, we will show that our method is also scalable.

\subsubsection{Real-World Datasets} \label{sec:rw_ds}

\noindent \textbf{California housing dataset.} 
In this section, we use our method to explain the well-adopted California housing dataset. In Figure \ref{fig:housing}, we observe that the longitude and latitude combine synergistically. Alone, neither the longitude nor the latitude have great predictive power of housing prices. On the contrary, when considered together, these variables reveal an exact location and become a very useful indicator. Another interesting feature of these results is that the ocean proximity is only redundant with respect to the longitude, as the coast is to the east. Other methods are unable to reveal such relationships. For extra experiments on classic ML datasets, please refer to Appendix \ref{app:extra_expts}. 

\textbf{BRCA dataset.}
We now demonstrate PIDF's ability to describe the relationships between genes in the BRCA dataset (available here: \url{https://portal.gdc.cancer.gov/projects/TCGA-BRCA}) \citep{tomczak2015review,berger2018}. This dataset details the RNA expression levels of many genes in patients with and without breast cancer. Of these genes, 10 are known chemically to cause cancer. Previously, those using this dataset to study feature importance \citep{catav21,janssen2023} have aimed to develop methods that identify these 10 as the most important. In this study, we instead focus on how these 10 genes interact to cause breast cancer. Figure \ref{fig:brca} reveals that CDK6 interacts synergistically with multiple genes. This is because CDK6 regulates gene expression levels in the cell, which promotes or inhibits division \citep{nebenfuehr2020role,goel2022targeting}. Consequently, in a healthy cell CDK6 expression levels are correlated with those of the genes it regulates. On the contrary, this is not the case in cancerous cells \citep{malumbres2009cell}, where the CDK6 present becomes overactive, leading to expression levels that exceed what is expected. PIDF, unlike the baselines, indicates that we require the expression levels of both CDK6 and the genes it regulates to predict cancer, which is consistent with the overactive CDK6 hypothesis.\ref{app:neurons_sampling}.\begin{figure}[h]
\begin{center}
\centerline{\includegraphics[width=\columnwidth]{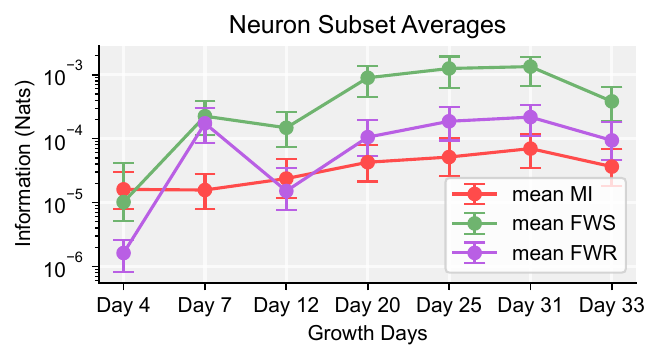}}
\caption{Average FWR, FWS, and MI per neuron as synaptic connections are re-established.}
\label{fig:neurons} 
\end{center}
\vskip -0.3in
\end{figure}

\textbf{Neuron growth experiment}. In this experiment, we study the information-theoretic quantities that characterize the spiking interactions that form within a dissociated neural culture over 33 days. Full details of the dataset are available in Appendix \ref{app:neurons_sampling}.
We now discuss the results presented in Figure \ref{fig:neurons}, where we observe the average FWR, FWS and MI for each neuron at each day over different neuron subsets. An initial increase in redundancy is followed by a decrease. This phenomenon aligns with the principles of Hebbian theory \citep{hebb05}, where the initial surge in redundancy is attributed to the exploratory phase of network formation, during which connections are established randomly. Subsequently, synaptic links that do not exhibit synchronized firing patterns are pruned. This adaptive reconfiguration results in not only a reduction in the observed redundancy, but also a gradual increase in synergy and mutual information.

\subsection{PIDF for Feature Selection}
In Section \ref{sec:fs}, we developed a simple algorithm based on PIDF for feature selection. We now evaluate this technique. 
Each synthetic dataset in Section \ref{sec:novsyn_expts}, characterized by certain relationships, has an associated set of optimal features.  In this section, we verify our method selects this optimal subset. 
We adopt the methods described in \citet{wollstadt23}, \citet{westphal24} (TERC), and \citet{breiman2001} (PI) as baselines.
In Table \ref{tab:1} we observe that PIDF selects the optimal number of features for all the datasets investigated; however, this is not the case for the baselines. The method developed in \citet{wollstadt23} fails to recognize fully synergistic variables as important (hence, the poor results achieved on the SVQ, TERC-1, and TERC-2 datasets). Meanwhile, TERC is not able to address the MSQ due to its inability to rank features based on their MCI.
\begin{table}[t]
\centering
\vskip -0.1in
\caption{Number of true positive (TP), false positive (FP), true negative (TN), and false negative (FN) features selected using each method. Red values represent optimal performance, and confidence intervals were omitted since negligible variations were observed.} 
\label{tab:1}
\scriptsize
\setlength{\tabcolsep}{0.65pt} 
\begin{tabular}{c|cccc|cccc|cccc|cccc}
\multirow{2}{*}{Dataset} & \multicolumn{4}{c|}{\textbf{PIDF}} & \multicolumn{4}{c|}{\textbf{Wollstadt}} & \multicolumn{4}{c|}{\textbf{TERC}} & \multicolumn{4}{c}{\textbf{PI}} \\
& TP & FP & TN & FN & TP & FP & TN & FN & TP & FP & TN & FN & TP & FP & TN & FN \\
\hline
RVQ & \textcolor{red}{2} & \textcolor{red}{0} & \textcolor{red}{1} & \textcolor{red}{0} & \textcolor{red}{2} & \textcolor{red}{0} & \textcolor{red}{1} & \textcolor{red}{0} & \textcolor{red}{2} & \textcolor{red}{0} & \textcolor{red}{1} & \textcolor{red}{0} & \textcolor{red}{2} & \textcolor{red}{0} & \textcolor{red}{1} & \textcolor{red}{0} \\
\hline
SVQ & \textcolor{red}{2} & \textcolor{red}{0} & \textcolor{red}{0} & \textcolor{red}{0} & 0 & 0 & 0 & 2 & \textcolor{red}{2} & \textcolor{red}{0} & \textcolor{red}{0} & \textcolor{red}{0} & \textcolor{red}{2} & \textcolor{red}{0} & \textcolor{red}{0} & \textcolor{red}{0} \\
\hline
MSQ & \textcolor{red}{1} & \textcolor{red}{0} & \textcolor{red}{2} & \textcolor{red}{0} & \textcolor{red}{1} & \textcolor{red}{0} & \textcolor{red}{2} & \textcolor{red}{0} & 0 & 2 & 0 & 1 & 1 & 2 & 0 & 0 \\
\hline
WT & \textcolor{red}{2} & \textcolor{red}{0} & \textcolor{red}{1} & \textcolor{red}{0} & \textcolor{red}{2} & \textcolor{red}{0} & \textcolor{red}{1} & \textcolor{red}{0} & \textcolor{red}{2} & \textcolor{red}{0} & \textcolor{red}{1} & \textcolor{red}{0} & \textcolor{red}{2} & \textcolor{red}{0} & \textcolor{red}{1} & \textcolor{red}{0} \\
\hline
TERC-1 & \textcolor{red}{3} & \textcolor{red}{0} & \textcolor{red}{3} & \textcolor{red}{0} & 0 & 0 & 3 & 3 & \textcolor{red}{3} & \textcolor{red}{0} & \textcolor{red}{3} & \textcolor{red}{0} & 2 & 0 & 3 & 1 \\
\hline
TERC-2 & \textcolor{red}{3} & \textcolor{red}{0} & \textcolor{red}{3} & \textcolor{red}{0} & 0 & 0 & 3 & 3 & \textcolor{red}{3} & \textcolor{red}{0} & \textcolor{red}{3} & \textcolor{red}{0} & 3 & 3 & 0 & 0 \\
\hline
UMFIBR & \textcolor{red}{1} & \textcolor{red}{0} & \textcolor{red}{3} & \textcolor{red}{0} & \textcolor{red}{1} & \textcolor{red}{0} & \textcolor{red}{3} & \textcolor{red}{0} & \textcolor{red}{1} & \textcolor{red}{0} & \textcolor{red}{3} & \textcolor{red}{0} & 1 & 3 & 0 & 0 \\
\hline
SG & \textcolor{red}{3} & \textcolor{red}{0} & \textcolor{red}{0} & \textcolor{red}{0} & \textcolor{red}{3} & \textcolor{red}{0} & \textcolor{red}{0} & \textcolor{red}{0} & \textcolor{red}{3} & \textcolor{red}{0} & \textcolor{red}{0} & \textcolor{red}{0} & 2 & 0 & 0 & 1 \\
\end{tabular}
\vskip -0.1in
\end{table}

\section{CONCLUDING REMARKS}
\textbf{Summary of the Contributions.} In this paper, we have presented PIDF, a novel paradigm for simultaneous data interpretability and feature selection based on per-feature synergy, redundancy and mutual information.
Through an extensive evaluation using both synthetic and real-world data, 
we have demonstrated its effectiveness in analyzing and interpreting systems characterized by complex interactions.
As part of our future agenda, we plan to explore the potential of applying PIDF to open problems in a variety of scientific fields. 

\textbf{Limitations.} The main limitation of PIDF is its temporal complexity, which scales as a function $\mathcal{O}(k^2)$, where $k$ is the number of features. It is worth noting that it is not directly comparable with the other methods since it captures feature-wise synergistic interactions, an intrinsically computationally expensive task. Consequently, in order to scale-up to very-large feature spaces, it is necessary to adopt hierarchical solutions for the comparison of sets of variables in PIDF.


\bibliography{neurips}

\section*{Checklist}

 \begin{enumerate}

 \item For all models and algorithms presented, check if you include:
 \begin{enumerate}
   \item A clear description of the mathematical setting, assumptions, algorithm, and/or model. [Yes]
   \item An analysis of the properties and complexity (time, space, sample size) of any algorithm. [Yes]
   \item (Optional) Anonymized source code, with specification of all dependencies, including external libraries. [Not Applicable]
 \end{enumerate}

 \item For any theoretical claim, check if you include:
 \begin{enumerate}
   \item Statements of the full set of assumptions of all theoretical results. [Yes]
   \item Complete proofs of all theoretical results. [Yes]
   \item Clear explanations of any assumptions. [Yes]     
 \end{enumerate}

 \item For all figures and tables that present empirical results, check if you include:
 \begin{enumerate}
   \item The code, data, and instructions needed to reproduce the main experimental results (either in the supplemental material or as a URL). [Yes]
   \item All the training details (e.g., data splits, hyperparameters, how they were chosen). [Yes]
         \item A clear definition of the specific measure or statistics and error bars (e.g., with respect to the random seed after running experiments multiple times). [Yes]
         \item A description of the computing infrastructure used. (e.g., type of GPUs, internal cluster, or cloud provider). [Yes]
 \end{enumerate}

 \item If you are using existing assets (e.g., code, data, models) or curating/releasing new assets, check if you include:
 \begin{enumerate}
   \item Citations of the creator If your work uses existing assets. [Yes]
   \item The license information of the assets, if applicable. [Not Applicable]
   \item New assets either in the supplemental material or as a URL, if applicable. [Yes]
   \item Information about consent from data providers/curators. [Not Applicable]
   \item Discussion of sensible content if applicable, e.g., personally identifiable information or offensive content. [Not Applicable]
 \end{enumerate}

 \item If you used crowdsourcing or conducted research with human subjects, check if you include:
 \begin{enumerate}
   \item The full text of instructions given to participants and screenshots. [Not Applicable]
   \item Descriptions of potential participant risks, with links to Institutional Review Board (IRB) approvals if applicable. [Not Applicable]
   \item The estimated hourly wage paid to participants and the total amount spent on participant compensation. [Not Applicable]
 \end{enumerate}

 \end{enumerate}

\newpage
\appendix
\onecolumn
\section{In Depth Comparison with PID}
\label{app:comarison}
In Section \ref{sec:rel_work}, we motivate PIDF by comparing it to existing feature importance and selection methods. However, we were unable to do an in-depth comparison of PIDF to pre-existing methods of calculating synergy and redundancy. Consequently, in this section, we provide a more thorough comparison. 

In \citet{williams2010}, the authors established a framework that identifies all possible interactions among features predicting a target, which has since been applied across multiple fields, including machine learning \citep{liang2023quantifying}, neuroscience \citep{luppi2024information}, complexity studies \citep{varley2022emergence}, cellular biology \citep{cang2020inferring}, and physics \citep{rosas2022disentangling}. In this paper, we aim to develop a method that allows us to understand with what other features and to what extent does each feature interact synergistically and redundantly. One would expect PID may be the solution. However, in its current form it comes with the following associated drawbacks: first, calculating the terms is intractable, while estimating them is subject to convergence issues and size limitations \citep{makkeh2018broja,makkeh2019maxent3d_pid,pakman2021}. Second, the number of terms needed to describe the system equals the $n-1$'th Dedekind number, where $n$ is the number of features. This number is impractically large, consider that a system with nine variables would require approximately $5 \times 10^{22}$ terms, while for ten variables, the Dedekind number remains unknown. \citet{varley2022emergence} reduced the number of investigable quantities by averaging the contribution of ‘layers’ in the PID lattice. The different layers can be thought of as representing different levels of redundancy or synergy. Consequently, through this spectral per-layer view one comes to understand how overall synergistic or redundant the system is. However, despite these simplifications, the calculations remained too complex for applications involving more than a few features. O-information \citep{rosas2019quantifying}, II \citep{chechik2001group}, correlational importance \citep{nirenberg2001retinal} and synergistic mutual information \citep{griffith2015} can all be used to estimate the synergy or redundancy of large sets of variables. However, they are unable to consider individual features. The calculation of such quantities does not assist us in answering the question: does a feature interact synergistically, or redundantly, and with what? Instead, they reveal emergent properties of sets. In this work we overcome such issues, by adopting the feature-wise perspective introduced in \citet{wollstadt23}. Specifically, we re-write the II as a two-body system, containing the feature of interest, as a single variable, and the remaining features as a second variable. Re-formulating the II in this way will allow us to develop measures of synergy and redundancy on a per-feature basis (labeled FWS and FWR), which interact in a consistent way. Unlike its predecessors, this re-formulation will remain computable at scales beyond what is currently possible in PID. To emphasize the differences in practicality of these methods, we provide the following example. 

Let us suppose we have four features, $F_0$ and $F_1$ are random binary strings, while $F_0 = F_2$, $F_3 = F_1$, and $Y = F_0 + F_1$. To begin, let us outline the interactions in this system that should be present after we have decomposed the information:
\begin{itemize}
    \item $F_0$ is redundant with respect to $F_2$ and vice versa;
    \item $F_1$ is redundant with respect to $F_3$ and vice versa;
    \item $F_0$ or $F_2$ combines synergistically with $F_1$ or $F_3$ and vice versa. 
\end{itemize}
This is because the knowledge of one binary string alone only reduces the possible realizations of $Y$ by 1/3. For instance, if $f_0 = 1$ then $y = 1$ or $y = 2$, leading to only a $0.58$ bit reduction in uncertainty. On the other hand, having both $F_0$ and $F_1$ fully describes $Y$, leading to a $1.58$ bit reduction in uncertainty. Consequently, $I(Y;F_0,F_1)>I(Y;F_0)+I(Y;F_1)$ and our assumption is satisfied.

If we were to now attempt to decompose this information using the system developed in \citet{williams2010}, we would have to calculate $120$ terms. Moreover, no method currently exists with the ability to estimate these quantities. One can estimate the redundancy/synergy of the system by calculating its II. However, such methods fail to describe any properties of individual features. The only viable option is the PIDF algorithm. As a result, we guide the reader through the calculations for the feature $F_0$.

As stated, $I(Y;F_0) = 0.58$ bits. We now move onto the calculation of the FWS. The aim here is to isolate the subset of the existing features that maximize the synergistic contribution of $F_0$, in accordance with the definition of FWS. For our four features, we obtain the following:
\[
FWS(Y;F_0;\mathcal{F} \setminus F_0) = \max_{\mathcal{P}^{ms} \in \{\{ F_1 \} , \{ F_2 \} , \{ F_3 \} , \{ F_1, F_2 \}, \ldots\}} \left(I(Y;F_0, \mathcal{P}^{ms}) - I(Y; \mathcal{P}^{ms}) - I(Y; F_0) \right) .
\]
The result is that we have three subsets for which we always obtain the maximum: $\{F_1\}, \{F_3\}$ and $\{F_1, F_3\}$. The members of these subsets are therefore the features with which $F_0$ combines synergistically, where the final result is $ FWS(Y;F_0,\mathcal{F} \setminus F_0) = 0.42$. Due to redundancy among the synergistic features, there are three sets of maximum synergy, with the least cardinal sets being free of redundancy. Notably, our algorithm would only derive one of these subsets as the set of maximum synergy, which does lead to a slight loss in information. However, this comes with the associated speedups. Now, we calculate:
\begin{align*}
FWR(Y;F_0;\mathcal{F} \setminus F_0) &= FWS(Y;F_0;\mathcal{F}_{\backslash F_0})- I(Y; F_0; \mathcal{F}_{\backslash F_0}) \\
&= \max_{\mathcal{P}^{ms} \in \{\{ F_1 \} , \{ F_2 \} , \{ F_3 \} , \{ F_1, F_2 \}, \ldots\}} \left(I(Y;F_0, \mathcal{P}^{ms}) - I(Y; \mathcal{P}^{ms}) - I(Y; F_0) \right) \\
&\quad - \left(I(Y;F_0, F_1, F_2, F_3) - I(Y;F_0) - I(Y;F_1, F_2, F_3)\right).
\end{align*}
From our synergy calculations, we know this leads to:
\[
\left( I(Y;F_0, F_1) -I(Y;F_0) - I(Y; F_1) \right)- \left(I(Y;F_0, F_1, F_2, F_3) - I(Y;F_0) - I(Y;F_1, F_2, F_3)\right) = 0.58 + 0.42,
\]
implying that this binary feature is fully redundant. Overall, we have again demonstrated that our method can be used to illuminate synergistic and redundant contributions of individual features.

\section{Proofs}
\subsection{Proof of Theorem 1}\label{app:proof_tint}
\textit{Proof.} We now prove Theorem 1.
\begin{equation}
\begin{split}
\label{eqn:fwredtheor_proof}
MCI(Y;F_i;\mathcal{F}_{\setminus F_i}) - FWR(Y;F_i;\mathcal{F}_{\setminus F_i}) =&\\
= &I(Y;F_i) + FWS(Y;F_i;\mathcal{F}_{\setminus F_i}) - FWR(Y;F_i;\mathcal{F}_{\setminus F_i}) \\ 
& \text{(Via our definition of MCI)} \\ 
= & I(Y;F_i) + I(Y;F_i;\mathcal{F}_{\setminus F_i})\\
& \text{(Via our definition of FWR and cancelling FWS)} \\ 
= & I(Y;F_i) + I(Y;F_i,\mathcal{F}_{\setminus F_i}) - I(Y;F_i) - I(Y;\mathcal{F}_{\setminus F_i})\\
= & I(Y;F_i,\mathcal{F}_{\setminus F_i}) - I(Y;\mathcal{F}_{\setminus F_i})
\end{split}
\end{equation}

\subsection{Proof of Theorem 2}\label{app:proof_t2}
\textit{Proof.} We begin by proving the lower bound. This proof shares the first few steps with those described in \citet{steudel15,lau23,janssen2023}, before we then use our assumption to complete the proof. 
\begin{equation}
\begin{split}
\label{eqn:t2_up}
I(Y;F_i,\mathcal{P}_{\backslash \{F_i, F_j\}}, F_j) &- I(Y;\mathcal{P}_{\backslash  \{F_i, F_j\}}, F_j) \\  =&I(Y;F_i|\mathcal{P}_{\backslash  \{F_i, F_j\}}, F_j) + I(Y;\mathcal{P}_{\backslash  \{F_i, F_j\}}, F_j) - I(Y;\mathcal{P}_{\backslash  \{F_i, F_j\}}, F_j) \\ & \text{(via the chain rule)} \\  =& I(F_i;Y|\mathcal{P}_{\backslash  \{F_i, F_j\}}, F_j) \\ & \text{(via symmetry)} \\  =& I(F_i;Y,\mathcal{P}_{\backslash  \{F_i, F_j\}}, F_j) - I(F_i;\mathcal{P}_{\backslash  \{F_i, F_j\}}, F_j)  \\ & \text{(via the chain rule)} \\  \geq & I(F_i;Y,\mathcal{P}_{\backslash  \{F_i, F_j\}}) - I(F_i,\mathcal{P}_{\backslash  \{F_i, F_j\}}) - I(F_i,F_j)  \\ & \text{(via the monotonicity of MI and our assumption)} \\  \geq & I(F_i;Y|\mathcal{P}_{\backslash  \{F_i, F_j\}}) + I(F_i;\mathcal{P}_{\backslash  \{F_i, F_j\}}) - I(F_i,\mathcal{P}_{\backslash  \{F_i, F_j\}}) - I(F_i,F_j) \\ & \text{(via the chain rule)} \\
 \geq & I(Y;F_i|\mathcal{P}_{\backslash  \{F_i, F_j\}}) - I(F_i,F_j) \\ & \text{(via symmetry)} \\
 \geq & I(Y;F_i,\mathcal{P}_{\backslash  \{F_i, F_j\}}) - I(Y;\mathcal{P}_{\backslash  \{F_i, F_j\}}) - I(F_i,F_j) \\ & \text{(via chain rule)} \\ \theta(Y;F_i;F_j;\mathcal{P}_{\backslash \{F_i, F_j\}}) \geq &  - I(F_i,F_j)
\end{split}
\end{equation} 
We now provide the upper bound to complete the proof.
\begin{equation}
\begin{split}
\label{eqn:t2_lb}
 I(Y;F_i,\mathcal{P}_{\backslash \{F_i, F_j\}}, F_j)& - I(Y;\mathcal{P}_{\backslash \{F_i, F_j\}}, F_j)  \\=&I(Y;F_i|\mathcal{P}_{\backslash \{F_i, F_j\}}, F_j) + I(Y;\mathcal{P}_{\backslash \{F_i, F_j\}}, F_j) - I(Y;\mathcal{P}_{\backslash \{F_i, F_j\}}, F_j) \\& \text{(via the chain rule)} \\  =& I(F_i;Y|\mathcal{P}_{\backslash \{F_i, F_j\}}, F_j) \quad \text{(via symmetry)} \\  =& I(F_i;Y,\mathcal{P}_{\backslash \{F_i, F_j\}}, F_j) - I(F_i;\mathcal{P}_{\backslash \{F_i, F_j\}}, F_j)  \quad \text{(via the chain rule)} \\  \leq & I(F_i;Y,\mathcal{P}_{\backslash \{F_i, F_j\}}, F_j) - I(F_i; F_j)  \quad \text{(via the monotonicity of MI)} \\  \leq & I(F_i;Y,F_j|\mathcal{P}_{\backslash \{F_i, F_j\}}) +I(F_i;Y,\mathcal{P}_{\backslash \{F_i, F_j\}})  - I(F_i; F_j) \\& \text{(via the chain rule)} \\  \leq & I(F_i;Y,F_j|\mathcal{P}_{\backslash \{F_i, F_j\}}) +I(F_i;Y|\mathcal{P}_{\backslash \{F_i, F_j\}}) +I(F_i;\mathcal{P}_{\backslash \{F_i, F_j\}})  - I(F_i; F_j) \\& \text{(via the chain rule)} \\  \leq & I(F_i;Y,F_j|\mathcal{P}_{\backslash \{F_i, F_j\}}) +I(Y;F_i|\mathcal{P}_{\backslash \{F_i, F_j\}}) +I(F_i;\mathcal{P}_{\backslash \{F_i, F_j\}})  - I(F_i; F_j) \\& \text{(via symmetry)} \\ \leq & I(F_i;Y,F_j|\mathcal{P}_{\backslash \{F_i, F_j\}}) + I(Y;F_i,\mathcal{P}_{\backslash \{F_i, F_j\}}) \\& - I(Y;\mathcal{P}_{\backslash \{F_i, F_j\}}) +I(F_i;\mathcal{P}_{\backslash \{F_i, F_j\}})  - I(F_i; F_j) \\& \text{(via the chain rule)} \\ \leq & I(F_i;Y,F_j|\mathcal{P}_{\backslash\{F_i, F_j\}}) + \theta(Y;F_i,\mathcal{P}_{\backslash \{F_i, F_j\}})  +I(F_i;\mathcal{P}_{\backslash \{F_i, F_j\}})  - I(F_i; F_j) \\& \text{(via the chain rule)} \\ \theta(Y;F_i;F_j;\mathcal{P}_{\backslash \{F_i, F_j\}}) \leq & I(F_i;Y,F_j|\mathcal{P}_{\backslash \{F_i, F_j\}})  +I(F_i;\mathcal{P}_{\backslash \{F_i, F_j\}})  - I(F_i; F_j) \quad \\ \leq & 2H(F_i)  - I(F_i; F_j) \quad \text{(via the monotonicity of MI)} 
\end{split}
\end{equation} 
\subsection{Proof that under Assumption 1 FWR is a Linear Combination of Redundant Contributions.}\label{app:proof_fwr}
In Section \ref{sec:pip}, we explained that, for enhanced interpretability, FWR can be decomposed into individual redundant contributions from features. In this section, we provide formal evidence for this statement. 
First, we note that:
\begin{align*}
FWR(Y;F_i;\mathcal{F}_{\backslash F_i}) 
&= I(Y;F_i;\mathcal{P}^{ms}) \;-\; I(Y;F_i;\mathcal{F}_{\backslash F_i}) \\[6pt]
&= I\bigl(Y;F_i,\mathcal{P}^{ms}\bigr) - I\bigl(Y;F_i\bigr) \;-\; I\bigl(Y;\mathcal{P}^{ms}\bigr)\\
&\quad\quad - \bigl( I\bigl(Y;F_i;\mathcal{F}_{\backslash F_i}\bigr) - I\bigl(Y;F_i\bigr) - I\bigl(Y;\mathcal{F}_{\backslash F_i}\bigr) \bigr) \\[6pt]
&= I\bigl(Y;F_i,\mathcal{P}^{ms}\bigr) \;-\; I\bigl(Y;\mathcal{P}^{ms}\bigr) \;-\; I\bigl(Y;F_i;\mathcal{F}_{\backslash F_i}\bigr) \;+\; I\bigl(Y;\mathcal{F}_{\backslash F_i}\bigr).
\end{align*}

\noindent
Next, consider the upper bounds of FWR. Suppose $I\bigl(Y;\mathcal{P}^{ms}\bigr) = 0$, while all remaining terms in the expression above are equal to $H(Y)$. We obtain:
\begin{align*}
FWR(Y;F_i;\mathcal{F}_{\backslash F_i}) &\geq I\bigl(Y;F_i,\mathcal{P}^{ms}\bigr) \;-\; I\bigl(Y;F_i;\mathcal{F}_{\backslash F_i}\bigr) \;+\; I\bigl(Y;\mathcal{F}_{\backslash F_i}\bigr) \\[4pt]
&\geq H(Y)-\; H(Y) \;+\; I\bigl(Y;\mathcal{F}_{\backslash F_i}\bigr) \\[4pt]
&\geq I\bigl(Y;\mathcal{F}_{\backslash F_i}\bigr).
\end{align*}

\noindent
Because $F_i$ is fully redundant and maximally reduces the uncertainty of $Y$, we can further write:
\[
FWR(Y;F_i;\mathcal{F}_{\backslash F_i}) \;\geq\; I\bigl(F_i;\mathcal{F}_{\backslash F_i}\bigr).
\]

\noindent
Under our assumption, we can then ``un-roll'' the FWR into a sum of pairwise mutual information, completing the proof.

\section{Algorithms}\label{app:algs}

\subsection{Partial Information Decomposition of Features}
The full implementation of PIDF is presented in Algorithm \ref{algo_pidf}.

\begin{algorithm}[ht]
\label{alg:main}
    \caption{Partial Information Decomposition of Features}
    \label{alg:algorithmblah}
    \textbf{Input}: Sampled features $\mathcal{F}$, and target $Y$.\\
    \textbf{Output}: $(I(F_i;Y),FWS(Y;F_i;\mathcal{F}_{\setminus F_i}),CR(Y;F_i;\mathcal{F}_{\backslash F_i}),\mathcal{R}_{\backslash F_i} ) \forall F_i \in \mathcal{F}$. \\
    \begin{algorithmic} [1]
    \FOR{$F_i$ in $\mathcal{F}$}
    \STATE Calculate $I(F_i;Y)$
    \STATE $\mathcal{F}^{sorted}_{\backslash F_i} = \text{argsort}_{F_j \in \mathcal{F}_{\backslash F_i}} I(F_i;F_j)$
    \STATE Initialize $FWR(Y;F_i;\mathcal{F}_{\backslash F_j}) = 0$ and $\mathcal{R}_{\backslash F_i} = \emptyset$
    \FOR{$F_j$ in $\mathcal{F}^{sorted}_{\backslash F_i}$}
        \IF{$I(F_i,F_j) > 0$}
            \STATE $\mathcal{R}_{\backslash F_i} \cup F_j$
        \ENDIF
        \STATE Calculate $\theta(Y;F_i;F_j;\mathcal{F}^{sorted}_{\backslash \{F_i, F_j\}})$
        \IF{$\theta(Y;F_i;F_j,\mathcal{F}^{sorted}_{\backslash \{F_i, F_j\}}) \leq 0$} 
                \STATE $FWR(Y;F_i;\mathcal{F}^{sorted}_{\backslash F_j}) =  FWR(Y;F_i;\mathcal{F}^{sorted}_{\backslash F_j}) - \theta(Y;F_i;F_j;\mathcal{F}^{sorted}_{\backslash \{F_i, F_j\}})$
                \STATE $\mathcal{F}^{sorted}_{\backslash F_i} = \mathcal{F}^{sorted}_{\backslash \{ F_i, F_j\}}$.
        \ENDIF
    \ENDFOR
    \STATE  $FWS(Y;F_i;\mathcal{F}_{\setminus F_i}) = I(Y;F_i,\mathcal{F}_{\backslash F_i})- I(\mathcal{F}_{\backslash F_i};Y)- I(F_i;Y)$
    \STATE $\mathcal{P}^{ms} = \mathcal{F}^{sorted}_{\backslash F_i}$
    \RETURN $I(F_i;Y)$, $FWS(Y;F_i;\mathcal{F}_{\setminus F_i})$, $FWR(Y;F_i;\mathcal{F}_{\backslash F_j}), \mathcal{P}^{ms}, \mathcal{R}_{\backslash F_i} \quad \text{($F_i$'s redundant features)}$
    \ENDFOR
\end{algorithmic}
\label{algo_pidf}
\end{algorithm}

\subsection{PIDF for Feature Selection}
The algorithm for feature selection based on PIDF is presented in Algorithm
\ref{algo_pidf_feature_selection}.

\begin{algorithm}[ht]
    \caption{PIDF for Feature Selection}
    \label{alg:fs}
    \textbf{Input}: $(FWS(Y;F_i;\mathcal{F}_{\setminus F_i}), FWR(Y;F_i;\mathcal{F}_{\backslash F_i}), I(Y;F_i), \mathcal{R}_{\backslash F_i}) \forall F_i \in \mathcal{F}$ \\
    \textbf{Output}: $\mathcal{F}_*$ (a desirable feature set).\\ 
    \begin{algorithmic} [1]
    \STATE Initialize $\mathcal{F}_* = \{ \}$.
    \FOR{$F_i$ in $\mathcal{F}$}
    \IF{$FWS(Y;F_i;\mathcal{F}_{\setminus F_i})+I(Y;F_i)>FWR(Y;F_i;\mathcal{F}_{\backslash F_i})$}
    \STATE $\mathcal{F}_* \cup F_i$
    \ENDIF
    \ENDFOR
    \STATE $\mathcal{F}^{sorted} = \text{argsort}_{F_j \in \mathcal{F} \backslash \mathcal{F}_*} FWS(Y;F_i;\mathcal{F}_{\setminus F_i})+I(Y;F_i)$
    \FOR{$F_i$ in $\mathcal{F}^{sorted}$}
    \IF{$\mathcal{F}_* - \mathcal{R}_{\backslash F_i} = \mathcal{F}_*$}
    \STATE $\mathcal{F}_* \cup F_i$
    \ENDIF
    \ENDFOR
    \RETURN $\mathcal{F}_*$
\end{algorithmic}
\label{algo_pidf_feature_selection}
\end{algorithm}

\section{Example Calculations}\label{app:example_calc}
In this Appendix, we perform example calculations for the RVQ dataset. We first re-introduce the form of the data, before then explicitly calculating from lines 2-17 in Algorithm \ref{alg:main} for feature $F_0$. We only consider $F_0$ for space and clarity. We then calculate the results of Algorithm \ref{alg:fs} in full. The RVQ dataset is of the following form:
\begin{equation}
\begin{split}
\label{eqn:RVQ}
F_0 & \sim \text{Bernoulli}(\mu=0.5, \sigma=0.5)\\
F_1 & \sim \text{Bernoulli}(\mu=0.5, \sigma=0.5)\\
F_2 & \equiv F_1\\
 Y&= F_0 + 2F_1.
\end{split}
\end{equation}

\subsection{Partial Information Decomposition of Features Calculation}
\textbf{Line 2}: We begin by calculating $I(F_0;Y)$. Given that $y = f_0 + 2f_1$, where $f_1$ and $f_2$ can be either 0 or 1 (according to Equation \ref{eqn:RVQ}), $y$ can take the following values with equal chance $y \in \{0, 1, 2, 3\}$. It follows that $H(Y) = -\sum_{y \in A(Y)}p_Y(y)\log_e(p_Y(y)) = \sum_{y \in A(Y)} 0.25 \cdot 1.386 = 1.386 $. Where $A(\cdot)$ is an operator that produces all possible realizations of a random variable. Upon obtaining the value of $f_0$, we reduce $y$'s potential realizations by half. To explain this, let us suppose $f_0 = 0$. In this case, it must be true that $y = 0 \quad \text{or} \quad 2$, whereas if $f_0 = 1$ it then must be true that $y = 1 \quad \text{or} \quad 3$. Therefore, the following holds $H(Y|F_0) = -\sum_{y \in A(Y), f_0\in A(F_0)}p_{Y,F_0}(y|f_0)\log_e(p_{Y,F_0}(y|f_0)) = \sum_{y \in A(Y), f_0\in A(F_0)} 0.5 \cdot 0.693 = 0.693$. Given that $I(F_0;Y) = H(Y)-H(Y|F_0)$ it must be true that $I(F_0;Y) = 1.386 -0.693 = 0.693$. This result agrees with the MI presented in Figure \ref{fig:novsynth}.

\textbf{Lines 3-4}: These lines are simple for variable $F_0$ as $F_1$ and $F_2$ have been defined such that they are independent random processes; therefore, $I(F_0;F_1)=I(F_0,F_2)=0$ and the precise ordering of set $\mathcal{F}^{sorted}_{\backslash F_i}$ is unimportant. In this case, we let $\mathcal{F}^{sorted}_{\backslash F_i} = \{F_1, F_2 \}$. 

\textbf{Lines 5-14:} To begin we do not add $F_1$ to the set $\mathcal{R}_{\backslash F_i}$ as $I(F_0;F_1) = 0$. We now calculate $\theta(Y;F_0,F_1,\mathcal{F}_{\{F_0,F_1\}}) = (I(Y;F_0,\mathcal{F}_{\backslash F_0}) - I(Y;\mathcal{F}_{\backslash F_0})) - (I(Y;F_0,\mathcal{F}_{\backslash \{F_0, F_1\}}) - I(Y;\mathcal{F}_{\backslash \{F_0, F_1\}}))$. To begin, we calculate $I(Y;F_0,\mathcal{F}_{\backslash F_0}) = H(Y) = 1.386$, because the knowledge of variables $F_0,\mathcal{F}_{\backslash F_0}$ completely describes the target $Y$. Meanwhile, $I(Y;\mathcal{F}_{\backslash F_0}) = 0.693$, because without $F_0$ (via a calculation identical to that completed for $I(F_0;Y)$ above), we can only reduce the uncertainty of $Y$ by half. Therefore,  $(I(Y;F_0,\mathcal{F}_{\backslash F_0}) - I(Y;\mathcal{F}_{\backslash F_0})) = 0.693$. We now investigate the values of $I(Y;F_0,\mathcal{F}_{\backslash \{F_0, F_1\}})$ and $ I(Y;\mathcal{F}_{\backslash \{F_0, F_1\}})$. Let us first note that $F_1 \equiv F_2$; consequently, by removing $F_1$ from $\mathcal{F}$, we do not diminish the set's ability to reduce the uncertainty of $Y$. It follows that, $I(Y;\mathcal{F}_{\backslash F_0})=I(Y;\mathcal{F}_{\backslash \{F_0, F_1\}})$, and therefore $I(Y;F_0,\mathcal{F}_{\backslash \{F_0, F_1\}})- I(Y;\mathcal{F}_{\backslash \{F_0, F_1\}}) = 0.693$, where $\theta(Y;F_0,F_1,\mathcal{F}_{\{F_0,F_1\}}) = 0$. Consequently, $F_1$ remains in $\mathcal{F}^{sorted}_{\backslash F_0}$ according to line 12. 

We now repeat lines 4-14, but iterating from $F_1$ to $F_2$. Again, we do not add $F_2$ to the set $\mathcal{R}_{\backslash F_i}$ as $I(F_0;F_2) = 0$.We now move onto lines 9-14 by calculating $\theta(Y;F_0,F_2,\mathcal{F}^{sorted}_{\backslash \{ F_0, F_2 \}})$. We note that $F_2\equiv F_1$, as a result it must be true that $\theta(Y;F_0,F_2,\mathcal{F}_{\{F_0,F_2\}})=\theta(Y;F_0,F_1,\mathcal{F}_{\{F_0,F_1\}}) = 0$. Therefore. for feature $F_0$, we return $I(F_0;Y)=0.693, FWS(Y;F_0;\mathcal{F}_{\backslash F_0})=0, FWR(Y;F_0;\mathcal{F}_{\backslash F_0})=0, \mathcal{R}_{\backslash F_i} = \{\} $. 

Above we have calculated from lines 2-17 in Algorithm \ref{alg:main} for variable $F_0$ in the RVQ dataset. By applying the same procedure, one can obtain the following values for \( F_1 \): \( I(F_1;Y)=0.693 \), \( FWS(Y;F_1;\mathcal{F}_{\backslash F_1})=0 \), \( FWR(Y;F_1;\mathcal{F}_{\backslash F_1})=0.693 \), with \( \mathcal{R}_{\backslash F_i} = \{F_2\} \). Similarly, for \( F_2 \): \( I(F_2;Y)=0.693 \), \( FWS(Y;F_2;\mathcal{F}_{\backslash F_2})=0 \), \( FWR(Y;F_2,\mathcal{F}_{\backslash F_2})=0.693 \), and again \( \mathcal{R}_{\backslash F_i} = \{F_1\} \). 

\subsection{Feature Selection Calculation} 
We will now utilize the results obtained from the PIDF calculations in the previous section to select features. This will be done by explicitly carrying out the calculations as per Algorithm \ref{alg:fs}.

\textbf{Lines 1-6:} From the results generated in the last section, it is clear that$I(F_0;Y)+FWS(Y;F_0;\mathcal{F}_{\backslash F_0})- FWR(Y;F_0;\mathcal{F}_{\backslash F_0})=0.693$, and therefore $F_0$ is included in the set of selected features ($\mathcal{F}_* = \{F_0\}$). However, this is not the case for $F_1$ and $F_2$. For $F_1$ we calculate $I(F_1;Y)+FWS(Y;F_1;\mathcal{F}_{\backslash F_1})- FWR(Y;F_1;\mathcal{F}_{\backslash F_1})=0.693+0-0.693=0$. Similarly, for $F_2$ we get $I(F_2;Y)+FWS(Y;F_2;\mathcal{F}_{\backslash F_2})- FWR(Y;F_2;\mathcal{F}_{\backslash F_2})=0.693+0-0.693=0$. Consequently, neither $F_1$ or $F_2$ are added to our set of selected features at this stage. 

\textbf{Lines 7-12:} We begin by ordering the non-selected features by their MCI $I(F_i;Y)+FWS(Y;F_i,\mathcal{F})$ (line 7). However, in this case, we have $I(F_1;Y)+FWS(Y;F_1;\mathcal{F})=0.693$ and $I(F_2;Y)+FWS(Y;F_2;\mathcal{F})=0.693$; therefore, the order is not meaningful. To complete the calculation, we let $\mathcal{F}^{sorted} = \{F_1, F_2\}$, and apply line 9. Currently, we have $\mathcal{F}_* = \{F_0\}$ and $\mathcal{R}_{\backslash F_1} = \{F_2\}$, therefore $\mathcal{F}_* - \mathcal{R}_{\backslash F_1} = \mathcal{F}_*$ and $F_1$ is added to the set $\mathcal{F}_*$, such that $\mathcal{F}_* = \{ F_1, F_2\}$. Now repeating this step for $F_2$, we have $\mathcal{F}_* = \{F_0, F_1\}$ and $\mathcal{R}_{\backslash F_2} = \{F_1\}$, it is trivial to see that $\mathcal{F}_* - \mathcal{R}_{\backslash F_2} = \{F_0\} \neq \mathcal{F}_* $. As a result $F_2$ is not added to our final set of selected features, and our final result is $\mathcal{F}_* = \{F_0, F_1\}$.

\section{Hyperparameter Selection}\label{app:hyper}
For both the feature selection and feature interpretation experiments we used a simple grid search when deciding hyperparameters. PIDF, TERC and the method introduced by \citet{wollstadt23} all rely on the calculation of MI. As discussed in Section \ref{sec:pip}, we estimate MI across all methods using the 5 repetitions of the techniques described by \citet{belghazi2018}. For clarity regarding hyperparameters, we present the details of this technique in Algorithm \ref{alg:te_est}.
\begin{algorithm*}[ht]
    \caption{MI estimation.}
    \label{alg:te_est}
    \textbf{Input}: Training dataset $  ({f}^1, {y}^1, {f}^2, {y}^2... {f}^T, {y}^T) $\\
    \textbf{Output}:  $I(Y;F)$
    \begin{algorithmic}[1] 
        \STATE Initialize weights $\theta$. 
        \FOR{$1$ to $N$}
        \STATE Draw mini batch samples of length $b$ from the joint distribution of the actions and the state with all possible variables included $ p_{Y,F} \sim ((y^{1},f^{1}),\dots,(y^{b},f^{b}))$, and repeat for the marginal distribution $ p_{Y\otimes F} \sim ((y^{1},f^{r}),\dots,(y^{b},f^{r}))$ (where the superscript $r$ indicates random sampling).
        
        \STATE $I({Y};F) = \frac{1}{b}\sum^b F_{\theta}((y^{1},f^{1}),\dots,(y^{b},f^{b})) - \frac{1}{b}\sum^b\log{e^{F_{\theta}((y^{1},f^{r}),\dots,(y^{b},f^{r}))}}$
        \STATE $\theta \leftarrow \tilde{\nabla}_{\theta} I({Y};F)$
        \ENDFOR
        \RETURN $I({Y};{F})$
    \end{algorithmic}
    \caption{MINE \cite{belghazi2018}. }
    \label{alg:estimation}
\end{algorithm*}

\subsection{Explaining Data Hyperparameters}
For PIDF we selected $b=1000$, $N=20,000$ as defined in Algorithm \ref{alg:estimation}. Meanwhile, the learning rate of the network in use was $\alpha = 0.0001$ with an adam optimizer. Each network had one hidden layer consisting of 50 neurons. 
For UMFI and MCI, 100 trees were selected as optimal. Otherwise, hyperparameters were as described in \citet{janssen2023} and \citet{catav21} respectively. For PI, we also selected 100 random trees. 

\subsection{Feature Selection Hyperparameters}
For PIDF, TERC and the Wollstadt method we again selected $b=1000$, $N=20,000$ as defined in Algorithm \ref{alg:estimation}. Meanwhile, the learning rate of the network in use was $\alpha = 0.0001$. For PI, we again selected 100 random trees.

\section{Statistical Characterization of Synthetic Datasets}\label{app:synth_desc}

\noindent \textbf{Redundant variables question dataset (RVQ).}
This dataset is comprised of three stochastic binary arrays, $F_0,F_1,F_2$ of length $1000$, two of which are fully redundant with respect to one another. More formally, we have:
\begin{equation}
\begin{split}
\label{eqn:rvq_form}
F_0 & \sim \text{Bernoulli}(\mu=0.5, \sigma=0.5)\\
F_1 & \sim \text{Bernoulli}(\mu=0.5, \sigma=0.5)\\
F_2 & \equiv F_1\\
Y & = F_0 + 2*F_1.
\end{split}
\end{equation}

\noindent \textbf{Synergistic variables question dataset (SVQ).}
The features are two stochastic binary arrays, $F_0,F_1$ of length $1000$, where $Y=XOR(F_0, F_1)$ and $F_0 \neq F_1$. $F_0$ and $F_1$ have been defined such that they combine to provide synergistic information about the target. More formally:
\begin{equation}
\begin{split}
\label{eqn:svq_form}
F_0 & \sim \text{Bernoulli}(\mu=0.5, \sigma=0.5)\\
F_1 & \sim \text{Bernoulli}(\mu=0.5, \sigma=0.5)\\
Y & = XOR(F_0, F_1).
\end{split}
\end{equation}

\noindent \textbf{Multiple subsets question dataset (MSQ).}
The features are three stochastic binary arrays, $F_0,F_1,F_2$ of length $1000$; where, $Y = F_0 = F_1+F_2$. This can be re-written: \begin{equation}
\begin{split}
\label{eqn:msp_form}
F_1 & \sim \text{Bernoulli}(\mu=0.5, \sigma=0.5)\\
F_2 & \sim \text{Bernoulli}(\mu=0.5, \sigma=0.5)\\
Y & = F_0 = F_1+F_2.
\end{split}
\end{equation}

\noindent \textbf{Wollstadt toy dataset (WT).} The Wollstadt toy example is characterized by three features, formed by combining normal distributions:
\begin{equation}
\begin{split}
\label{eqn:wt_form}
\epsilon_1 & \sim \mathcal{N}(\mu=0, \sigma=1)\\
\epsilon_2 & \sim \mathcal{N}(\mu=0, \sigma=1)\\
\epsilon_3 & \sim \mathcal{N}(\mu=0, \sigma=1)\\
F_0 & = \epsilon_1 + 0.1 F_2\\
F_1 & = 0.8\epsilon_1 +0.2\epsilon_2 + 0.01*F_2\\
F_2 & \sim \mathcal{N}(\mu=0, \sigma=1) \\
Y & = sin(\epsilon_1) +0.1\epsilon_3.
\end{split}
\end{equation}

\noindent \textbf{TERC-1.} This dataset, introduced by \citet{westphal24}, contains six features, with the following characteristics: 
\begin{equation}
\begin{split}
\label{eqn:TERC-1}
F_0 & \sim \text{Bernoulli}(\mu=0.5, \sigma=0.5)\\
F_1 & \sim \text{Bernoulli}(\mu=0.5, \sigma=0.5)\\
F_2 & \sim \text{Bernoulli}(\mu=0.5, \sigma=0.5)\\
F_3 & \equiv F_0\\
F_4 & \equiv F_0\\
F_5 & \equiv F_0\\
 Y&= 
\begin{cases}
    0,& \text{if }f_1=f_2=f_2\\
    1,              & \text{otherwise}\\
\end{cases}
\end{split}
\end{equation}

\noindent \textbf{TERC-2.} This dataset, introduced by \citet{westphal24}, contains six features, with the following characteristics: 
\begin{equation}
\begin{split}
\label{eqn:TERC-2}
F_0 & \sim \text{Bernoulli}(\mu=0.5, \sigma=0.5)\\
F_1 & \sim \text{Bernoulli}(\mu=0.5, \sigma=0.5)\\
F_2 & \sim \text{Bernoulli}(\mu=0.5, \sigma=0.5)\\
F_3 & \equiv F_0\\
F_4 & \equiv F_1\\
F_5 & \equiv F_2\\
 Y&= 
\begin{cases}
    0,& \text{if }f_1=f_2=f_2\\
    1,              & \text{otherwise}\\
\end{cases}
\end{split}
\end{equation}

\noindent \textbf{UMFI blood relation dataset (UBR).} \citet{janssen2023} used the UBR dataset to demonstrate that their method had favorable properties in detecting blood relationships in causal graphs. This dataset is characterized by four features with the following characteristics. 
\begin{equation}
\begin{split}
\label{eqn:umfi_br}
\epsilon_1 & \sim \mathcal{U}(\mu=-1, \sigma=1)\\
\epsilon_2 & \sim \mathcal{U}(\mu=-0.5, \sigma=0.5)\\
\epsilon_3 & \sim \text{Exp}(1)\\
\epsilon_4 & \sim \mathcal{N}(\mu=0, \sigma=1)\\
F_0 & \sim \mathcal{N}(\mu=0, \sigma=1)\\
F_1 & = 3F_1 + \epsilon_1\\
F_2 & = \epsilon_4 + F_0\\
F_3 & = Y + \epsilon_3\\
 Y&= \epsilon_4 + \epsilon_2.
\end{split}
\end{equation}

\noindent \textbf{Synthetic genes dataset (SG).} Our final synthetic dataset was developed in \citet{anastassiou07} to demonstrate the types of relationships that could be present in genetic interactions. Formally $Y$ is:
\begin{equation}
\begin{split}
\label{eqn:sg}
Y\sim \text{Bernoulli}(\mu=0.5, \sigma=0.5),
\end{split}
\end{equation} 
where $y=1$ indicates cancer. The features meanwhile can be explained via the following bullet points:
\begin{itemize}
    \item In healthy samples, the probability that both $f_1=f_2=1$ is 95\%, whereas the remaining three joint states (00, 01 and 10) are equally likely. Meanwhile, in cancerous samples, the probability that  both $f_1=f_2=1$ is 5\%, whereas the remaining three joint states (00, 01 and10) remain equally likely.
    \item In  healthy samples, if $y=0$, then $F_3\sim \text{Bernoulli}(\mu=0.2, \sigma=0.4)$. Meanwhile, in cancerous samples, we have $F_3\sim \text{Bernoulli}(\mu=0.8, \sigma=0.4)$.
\end{itemize}
\section{Sampling Method for Neural Culture Data}\label{app:neurons_sampling}
\begin{figure*}[ht]
\vskip 0.2in
\begin{center}
\centerline{\includegraphics[width=\columnwidth]{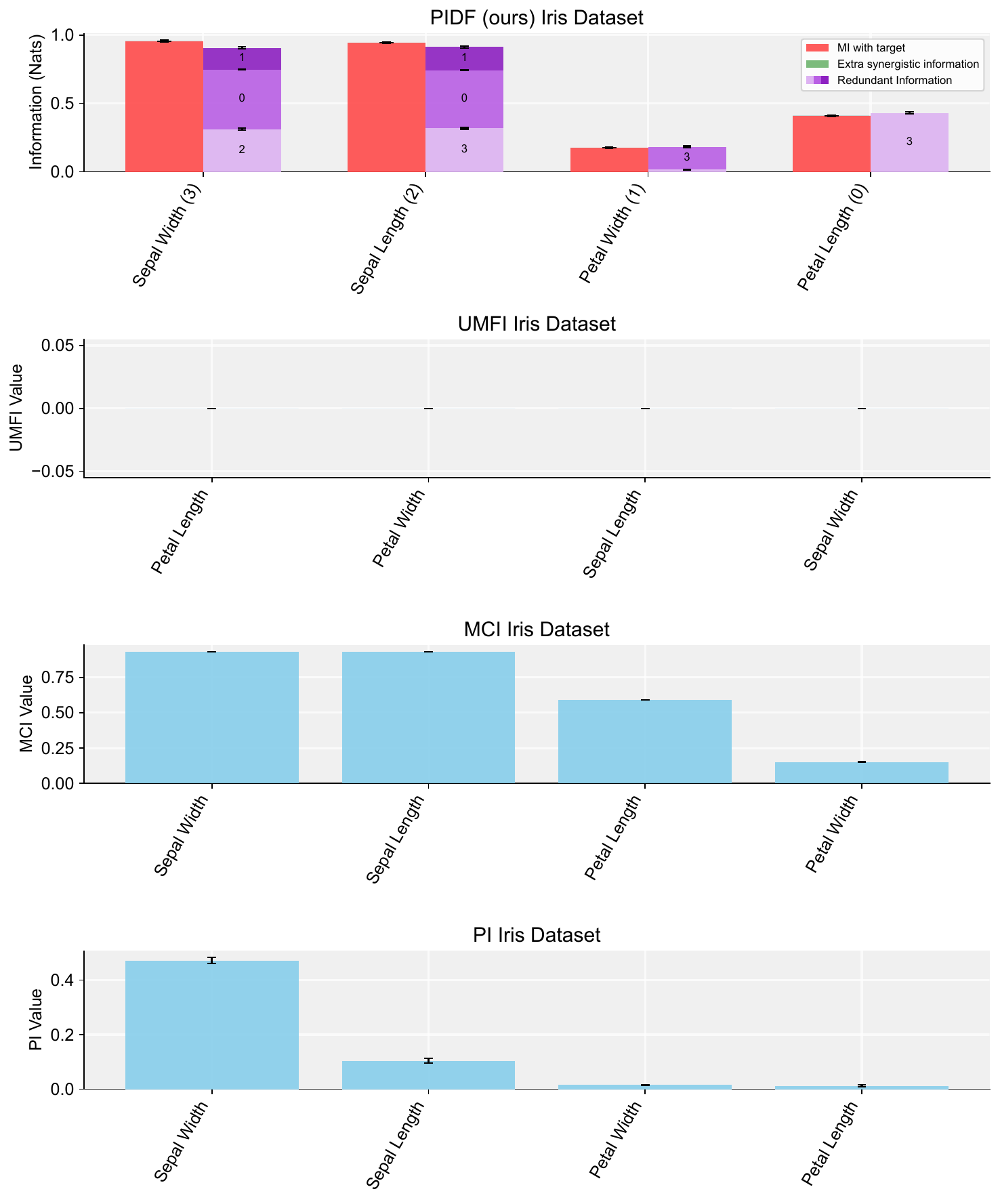}}
\caption{Comparison of feature importance methods using the classic Iris dataset.}
\label{fig:iris}
\end{center}
\vskip -0.2in
\end{figure*}
\begin{figure*}[ht]
\vskip 0.2in
\begin{center}
\centerline{\includegraphics[width=\columnwidth]{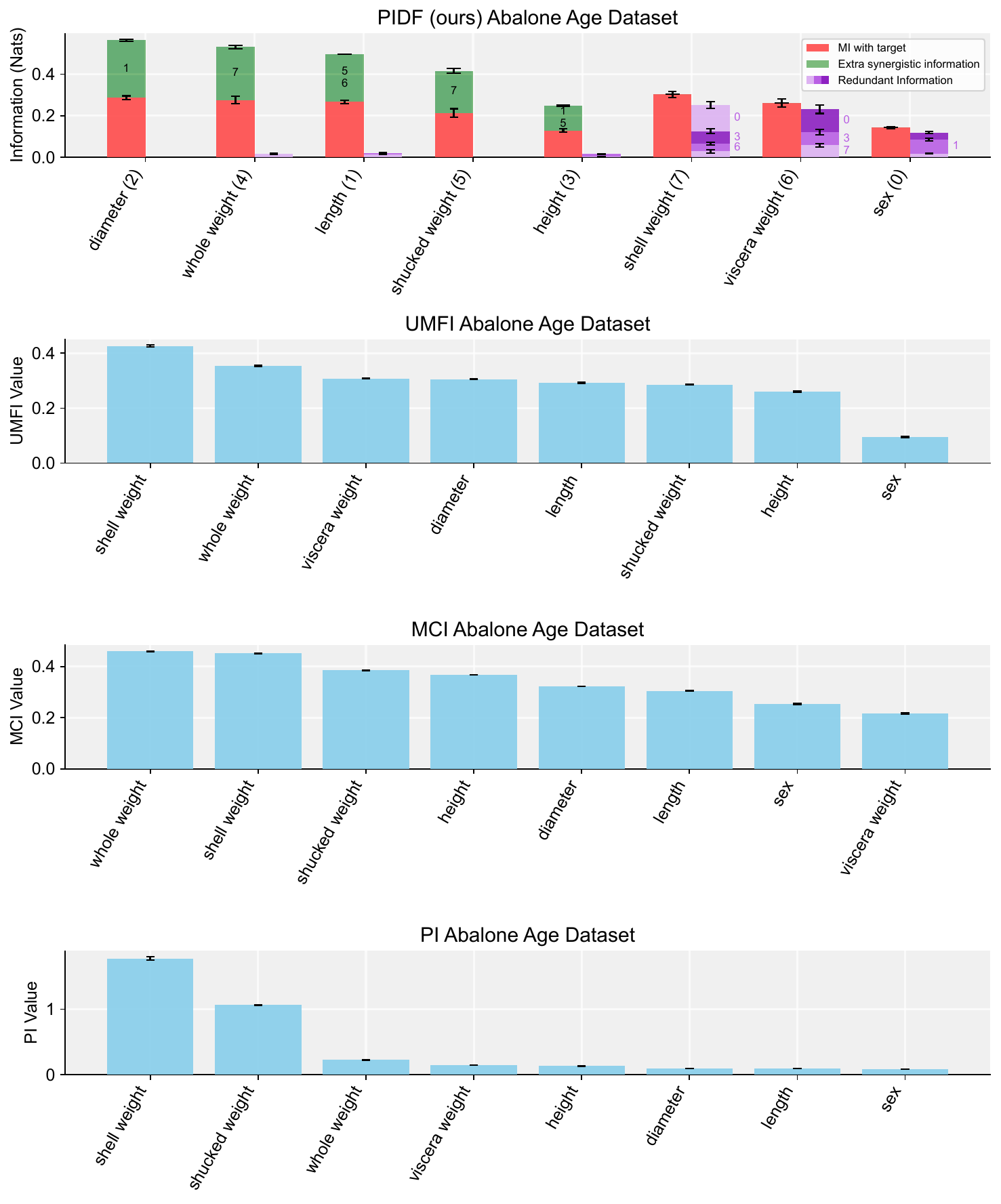}}
\caption{Comparison of feature importance methods using the classic Abalone dataset.}
\label{fig:Abalone}
\end{center}
\vskip -0.2in
\end{figure*}
\begin{figure*}[ht]
\vskip 0.2in
\begin{center}
\centerline{\includegraphics[width=\columnwidth]{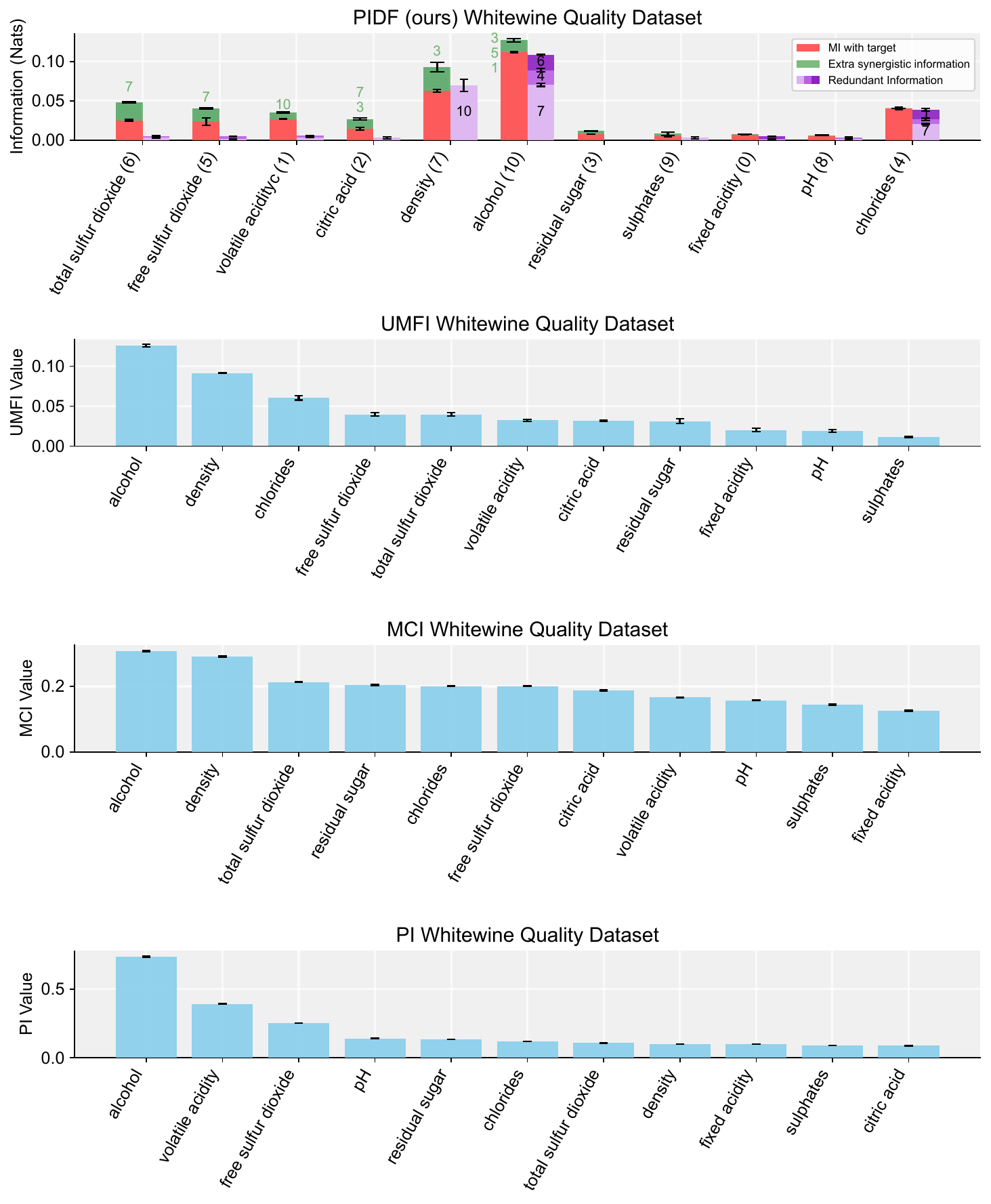}}
\caption{Comparison of feature importance methods using the classic whitewine quality dataset.}
\label{fig:Whitewine}
\end{center}
\vskip -0.2in
\end{figure*}
\begin{figure*}[ht]
\vskip 0.2in
\begin{center}
\centerline{\includegraphics[width=\columnwidth]{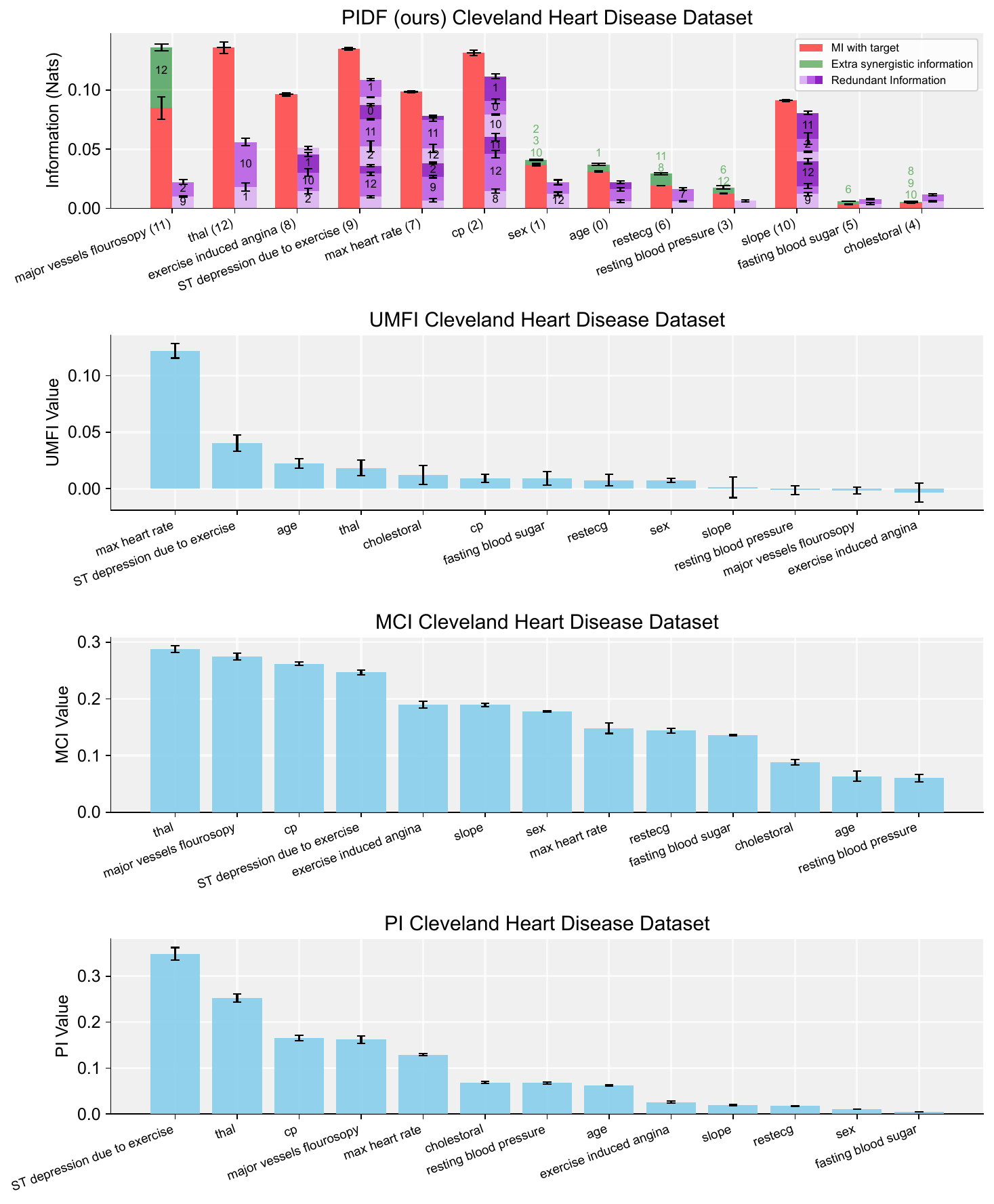}}
\caption{Cleveland heart disease prediction experiments.}
\label{fig:cleveland}
\end{center}
\vskip -0.2in
\end{figure*}
\begin{figure*}[ht]
\vskip 0.2in
\begin{center}
\centerline{\includegraphics[width=\columnwidth]{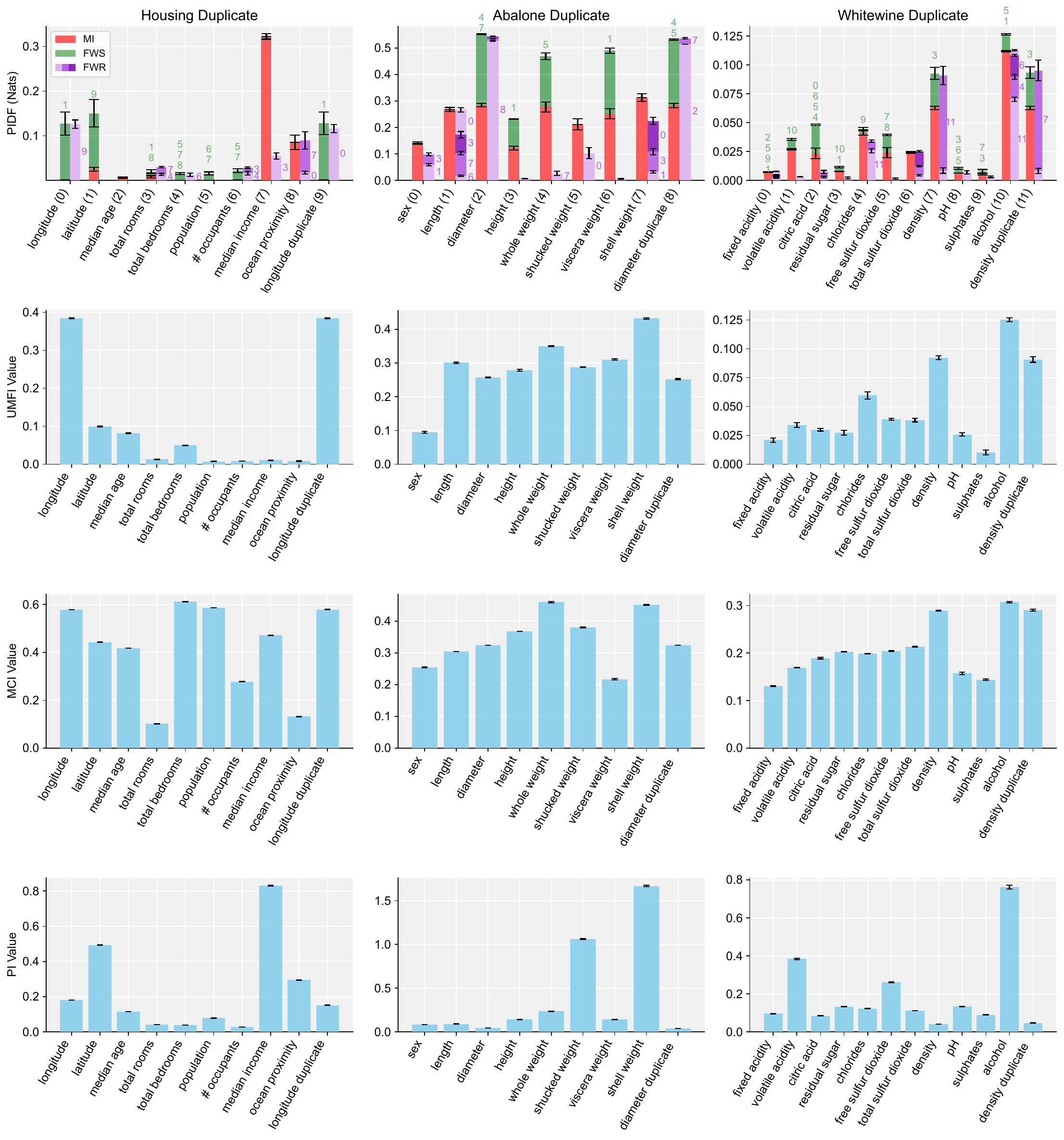}}
\caption{Feature duplication experiments. We apply PIDF to a modified version of the California housing, Abalone, and Whitewine datasets. These modified datasets are obtained by adding a duplicate feature to the original ones. The duplicate features we used in our experiments are longitude, diameter and density, respectively.}
\label{fig:duplicates}
\end{center}
\vskip -0.2in
\end{figure*}

We now provide more details regarding the firing neurons dataset, including the methods we used to sample from it. 
This dataset was introduced in \citet{wagenaar2006extremely} and is publicly accessible online\footnote{\url{ https://potterlab.gatech.edu/potter-lab-data-code-and-designs/}}. 
These datasets comprise multiunit spiking activities recorded from each of the 60 electrodes in the multielectrode array used to study the activations of dissociated neural cells. Specifically, the data from neural culture 2-2 were utilized. Comprehensive details regarding the cultivation and maintenance of these cultures are documented in \citet{wagenaar2006extremely}. The analysis focused on recordings from eight developmental stages of the culture, measured in days in vitro (DIV): 4, 7, 12, 16, 20, 25, 31, and 33. The recording from DIV 16 lasted for 60 minutes, while the recordings from the other days each lasted for 45 minutes. For the purposes of this analysis, the data were segmented into bins of 16 $\mu s$. The probability distributions required for calculating the three information-theoretic quantities of interest were generated by analyzing the spike trains from groups of ten distinct electrodes. Within each group, one electrode was randomly designated as the $Y$ variable, and the remainder as the features. For every time step in the spike trains, the states (spiking or not) of the feature electrodes at time $t$ and the $Y$ electrode at time $t+1$ were noted. This tabular data could then act as input to Algorithms \ref{alg:estimation} and \ref{alg:main}. This procedure was replicated for each set of non-identical electrodes, ensuring that groups with interchanged features were only considered once to prevent duplication. 
A full discussion of this dataset can be found in \citet{timme2018tutorial}.

\section{Further Experiments on Classic Machine Learning Datasets}\label{app:extra_expts}
\subsection{Iris Dataset}
In this section, we apply PIDF to the Iris dataset. One of the earliest and best-adopted datasets in ML, Iris was used to predict flower type given the petal length and width and sepal length and width.

In Figure \ref{fig:iris}, we observe that, unsurprisingly, there is a large amount of redundancy between the features in the Iris dataset. UMFI's pre-processing step, which removes redundancies between pairwise features, is unable to resolve the more complex redundancies we see Figure \ref{fig:iris}; hence, the assignment of feature importance scores of 0. Meanwhile, PIDF is the only method able to reveal the complex set of redundancies, showing that sepal width and sepal length are both well correlated with the target. But the sepal width also provides redundant information with respect to the petal length and width, making this the most informative feature, as revealed using PI.

\subsection{Abalone Dataset}
In this section, we apply PIDF to the Abalone dataset, which comprised information about the size and weight of an Abalone, and is used to predict its age.

In Figure \ref{fig:Abalone}, it is possible to observe that in this everyday dataset we see synergistic interactions occurring. The whole weight and shucked weight combine synergistically with the shell weight. Intuitively, this makes sense as finer grained detail about the weights of individual components of the animal may lead to greater knowledge of the animals age. Meanwhile, we observe that the viscera weight and shell weight are redundant with respect to one another, while this is not the case with whole and shucked weights. 

\subsection{Whitewine Dataset}
In this section, we apply PIDF to a classic ML dataset, which uses the chemical properties of wine to predict its quality.

In Figure \ref{fig:Whitewine}, we observe that the alcohol levels and density provide redundant information with respect to each other. This redundancy can be attributed to the fact that, apart from alcohol, wine primarily consists of water-based solubles. Consequently, the density of the wine is a function of the alcohol-water ratios. Interestingly, we see that the free and total sulfur dioxide features are not highly redundant with respect to one another. This is because, during the wine-making process, free sulfur dioxide is added for anti-microbial effects. Meanwhile, the bound sulfur dioxide is more dependent on the grape used.

\subsection{Cleveland Heart Disease Dataset}
In this section, we apply PIDF to a classic ML dataset, which uses medical data to predict cases of heart disease in Cleveland, Ohio.

In Figure \ref{fig:cleveland}, we observe that thalassemia intermedia and the number of major articles with blockages as detected by fluoroscopy combine synergistically to predict heart disease. 

\section{Feature Duplication Experiments}

In the main body of the paper, we demonstrated that our method could be used to successfully detect redundancies in synthetic data. Extending this validation, inspired by experiments in \citet{catav21}, we now demonstrate our method's ability to illuminate known redundancies in real-world datasets. This is achieved by duplicating features within the, widely recognized, California housing, Abalone, and Whitewine ML datasets. Subsequently, we provide evidence demonstrating the successful detection of these introduced redundancies by our methodology. These duplicated features will simply be added to the original datasets. 

In Figure \ref{fig:duplicates}, we showcase the outcomes from our experiments on duplicated features. The findings demonstrate that FWS and MI of the duplicated features mirror that of their non-duplicated counterparts. However, the information provided by these features is now redundant.

\section{Computational Requirements}
\label{app:computation}
In its current form, the execution of PIDF requires more computational resources that its other counterparts for feature interpretability \citep{catav21,janssen2023}. For example, executing the results for the TERC-1 and TERC-2 datasets, which contained six features each with 1000 instances took 45 minutes on a P100 NVIDIA GPU cluster. However, this can easily be reduced via the following methods:
\begin{enumerate}
\item As discussed in the conclusion, a hierarchical approach can be adopted.  Rather than removing individual redundant features, as per line 12 of Algorithm \ref{alg:main}, many can be checked and removed at once. 
\item Another way in which it is possible to scale the PIDF implementation  is by enhancing the efficiency of the MI estimation. Currently, we use MINE \citep{belghazi2018}, as the results are highly accurate and in Nats, which are easily interpretable. However, if one wishes just to compare the relative values of PIDF, without the need for values with units, more efficient methods can be used. In particular, we will make available a version of PIDF implemented using the method developed in \citet{covert2020}. By using this simple method of MI estimation,the time taken to execute PIDF is reduced by a factor of ten. However, the resulting values only represent relative importance and do not exactly correspond to the ground truth. We will discuss this method in detail in the following section.
\end{enumerate}

\section{Scalability Experiments}
In our experiments, we estimated MI using MINE. In this section, we discuss how MINE, despite its high accuracy, is computationally expensive. Therefore, it may be preferable to use a faster method, even if it sacrifices some accuracy. We now present an  implementation of PIDF using an alternative method of MI estimation. We will demonstrate that, using this approach, PIDF is highly scalable, making it applicable to real-world datasets, such as MNIST. We will discuss the inherent trade-offs between computation efficiency and accuracy of this alternative method.

\subsection{Comparison of MI Estimation Techniques}\label{app:mi_est}
\begin{figure}[h]
    \centering
    \includegraphics[width=\textwidth]{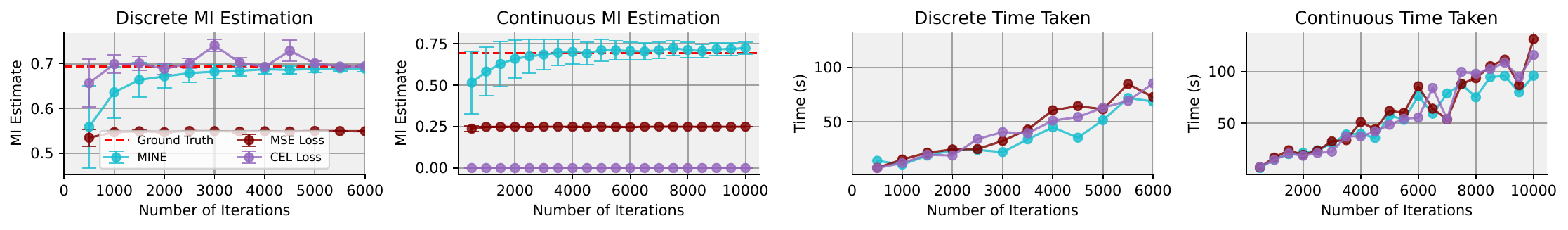}
    \caption{In these plots, we compare MINE with other MI estimators by illustrating the number of iterations required to converge on the ground truth MI. MINE takes the longest to converge but yields the most accurate estimates with minimal variation. CEL quickly produces accurate estimates on discrete data but exhibits high variation. An MSE loss converges rapidly but results in incorrect estimates.}
    \label{fig:mine_vs}
\end{figure}

First, we empirically represent the trade-off between efficiency and accuracy for three different MI estimation techniques \citep{covert2020}. In Figure \ref{fig:mine_vs}, we observe that MINE takes the longest to converge, but is the most accurate. Meanwhile, the method presented by \citet{covert2020} with an MSE loss converges rapidly, but it is not accurate.

\subsection{Alternative Method for the Calculation of MI and Trade-offs}\label{app:scalable_method}
Based on the results in the previous section, replacing MINE with the method presented by \citet{covert2020} with an MSE loss would likely result in a more efficient but less accurate approach. We briefly discuss the validity of such a method before presenting results. 

PIDF ideally would derive the values of FWR and FWS in bits; however, significant insights into the interactions can still be gained from the relative values of FWS, FWR, and MI. To clarify, Figure \ref{fig:mine_vs} illustrates the PIDF results obtained from two simple datasets using both MINE and our more efficient estimator. We observe that while the relative proportions of FWR, FWS, and MI remain consistent, the exact values in the two rightmost graphs are no longer accurate. Consequently, even when using the efficient estimator, it is still possible to discern the interactions and their respective elements.

\begin{figure}[h]
    \centering
    \includegraphics[width=\textwidth]{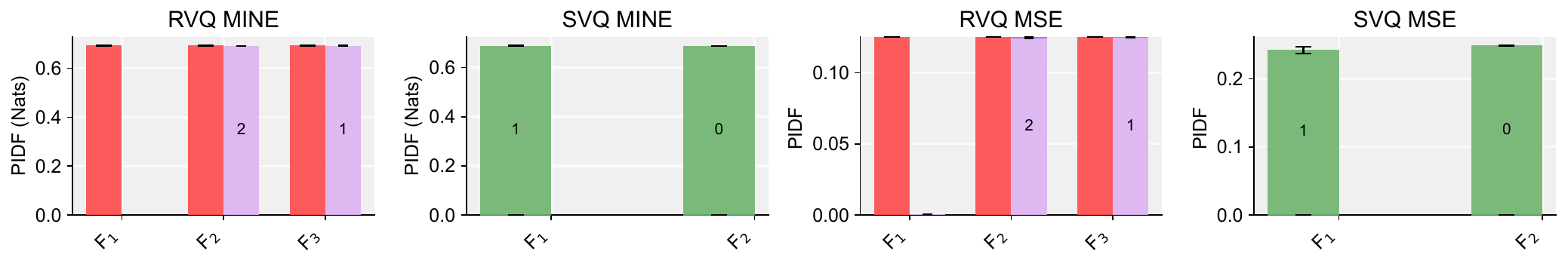}
    \caption{Comparison of the results of PIDF using MINE and PIDF using a different estimator.}
    \label{fig:4_subplots}
\end{figure}

\begin{figure}[h]
    \centering
    \includegraphics[width=\textwidth]{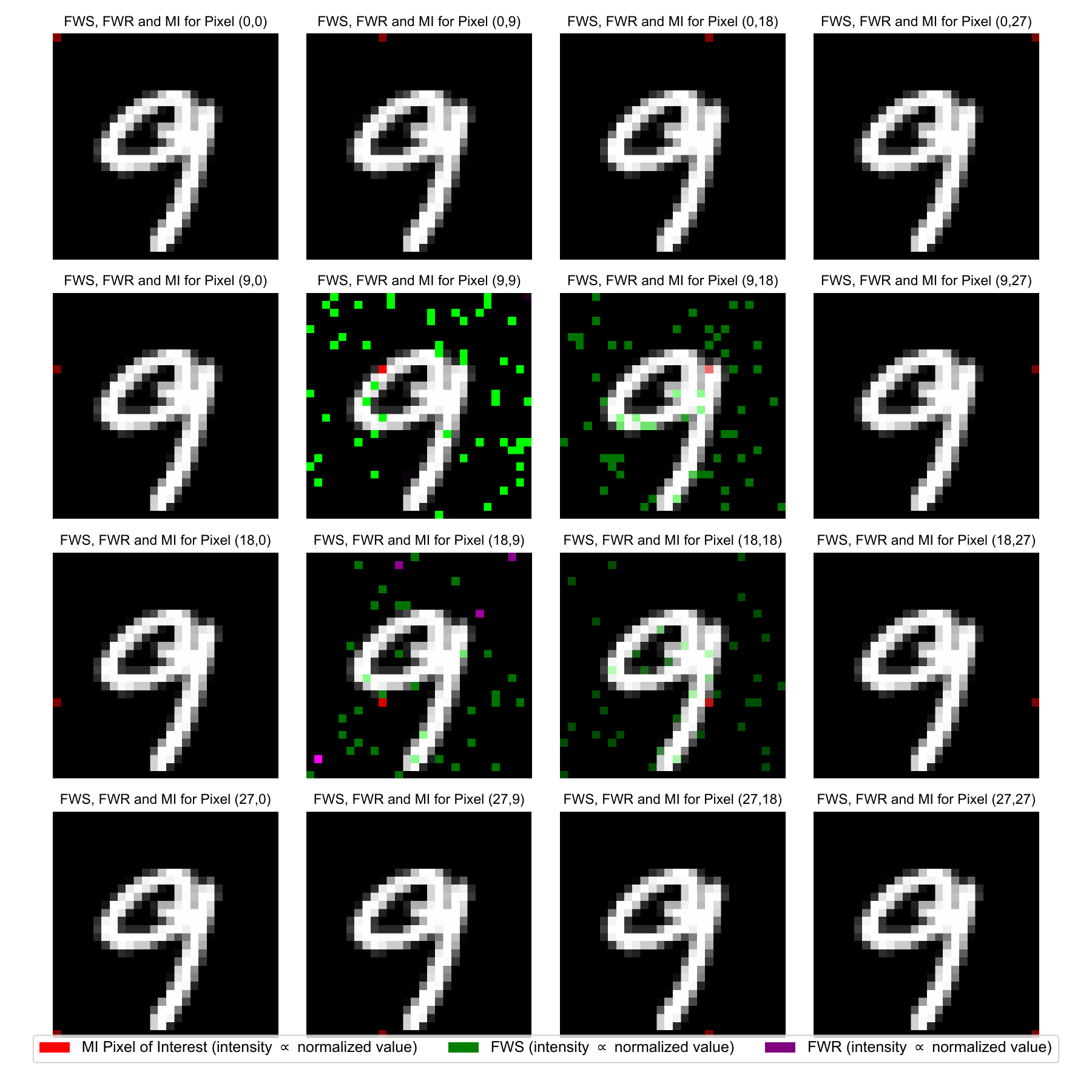}
    \caption{The results achieved when applying scalable PIDF to MNIST.}
    \label{fig:mnist_pidf}
\end{figure}

Consequently, we apply this more scalable method to the MNIST dataset and present the results in Figure \ref{fig:mnist_pidf}. The outer pixels in MNIST are often just black; consequently, only the central pixels contribute to strongly synergistic or redundant interactions, as they are the only pixels providing information. Furthermore, we observe a tendency for these pixels to interact synergistically. This occurs because, when considered together, their spatial relationships and patterns (such as edges, corners, and textures) emerge. These features are crucial for recognizing handwritten digits in MNIST. We might expect that pixels near one another would have high levels of redundancy. In fact, pixels close to one another did have high pairwise MIs. However, according to line 3 of Algorithm \ref{alg:main}, this caused them to be checked and removed first when identifying the set of maximum synergy. Consequently, other pixels were then identified as providing redundant information.

\end{document}